\documentclass{article}

\PassOptionsToPackage{numbers, compress}{natbib}
\usepackage[eandd, final]{neurips_2026}

\usepackage[utf8]{inputenc}
\usepackage[T1]{fontenc}
\usepackage{hyperref}
\usepackage{url}
\usepackage{booktabs}
\usepackage{amsfonts}
\usepackage{amsmath}
\usepackage{nicefrac}
\usepackage{microtype}
\usepackage{xcolor}
\usepackage{graphicx}
\usepackage{multirow}
\usepackage{caption}
\usepackage{subcaption}
\usepackage{amssymb}
\usepackage{placeins}
\newcommand{\cmark}{\checkmark}
\newcommand{\xmark}{--}

\title{WildBox: A Dataset and Benchmark for Aerial Monocular 3D Detection of African Savanna Wildlife}

\usepackage{authblk}  

\author[1,2]{\textbf{Vandita Shukla}}
\author[3]{\textbf{Kilian Meier}}
\author[4]{\textbf{Lucie Laporte-Devylder}}
\author[4]{\textbf{Camille Rondeau Saint-Jean}}
\author[5]{\textbf{Jenna M. Kline}}
\author[6,7,8]{\textbf{Blair R. Costelloe}}
\author[9]{\textbf{Devis Tuia}}
\author[1]{\textbf{Fabio Remondino}}
\author[2]{\textbf{Benjamin Risse}}

\affil[1]{3D Optical Metrology Unit, Fondazione Bruno Kessler, Trento, Italy}
\affil[2]{Computer Vision and Machine Learning Systems group, Institute for Geoinformatics, University of Muenster, Muenster, Germany}
\affil[3]{School of Civil, Aerospace and Design Engineering, University of Bristol, Bristol, UK}
\affil[4]{Department of Biology, University of Southern Denmark, Odense, Denmark}
\affil[5]{The Ohio State University, Columbus, Ohio, USA}
\affil[6]{Department of Collective Behavior,
Max Planck Institute of Animal Behavior
University of Konstanz, Konstanz, Germany}
\affil[7]{Centre for the Advanced Study of Collective Behaviour,
University of Konstanz, Konstanz, Germany}
\affil[8]{Department of Biology,
University of Konstanz, Konstanz, Germany}
\affil[9]{Environmental Computational Science and Earth Observation Laboratory, EPFL, Sion, Switzerland}

\begin{document}

\maketitle
\vspace{-3em}

\begin{center}
  \url{https://3dom-fbk.github.io/WildBox/}
\end{center}

\begin{figure}[h]
    \centering
    \includegraphics[width=1\linewidth]{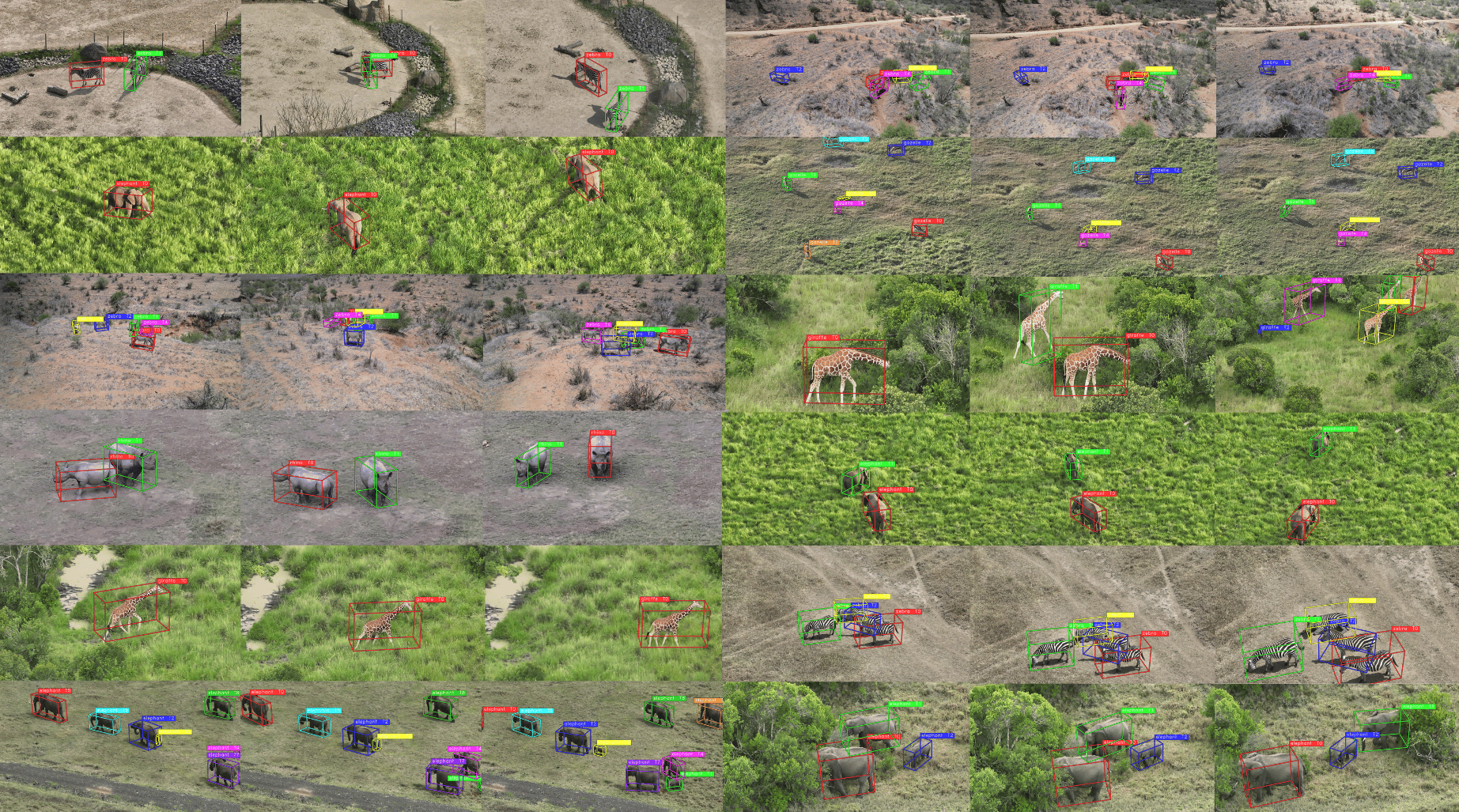}
    \caption{Sample annotations from our proposed \textbf{WildBox} dataset. Each row contains two three-frame strips, one per video sequence, with frames spaced across the sequence to show temporal coverage. The tracked 3D bounding boxes are drawn as coloured wireframes, with colour encoding instance ID consistently within each sequence. The dataset spans diverse taxa (zebras, elephants, gazelles, giraffes, rhinos), habitats, viewpoints, digital zoom, and group sizes captured from aerial drone footage.}
    \label{fig:data_overview}
\end{figure}

\begin{abstract}
We introduce WildBox, a dataset and benchmark for monocular 3D detection of wildlife from drone video, comprising 237{,}505 3D bounding box annotations across seven African savanna species grouped into six benchmark classes. 
Annotations follow a KITTI/Omni3D-compatible format in a per-segment scale-normalised camera frame, with instance identities maintained across each segment. 
We evaluate two open-vocabulary monocular 3D architectures, OVMono3D-LIFT and DetAny3D, under zero-shot, ground-truth 2D box prompt, and supervised fine-tuning protocols. 
Open-vocabulary 2D foundation models provide usable zero-shot wildlife localisation (50.55 AP@50), but zero-shot 3D detection collapses to 0.00 AP across both architectures and every 2D-input condition tested, including ground-truth 2D box prompts—isolating the failure to the 3D stage. 
Fine-tuning on WildBox recovers performance to 8.68$\pm$0.47 AP-BEV@0.50 and 13.17$\pm$0.69 AP3D macro. 
Depth contributes 84\% of normalised Hausdorff distance after fine-tuning and over 99\% in zero-shot, identifying monocular aerial depth as the dominant open problem in this regime. 
A coarse-to-fine curriculum—pretraining on a merged zebra class before fine-tuning on the Grévy's/plains split—improves macro 3D performance with less total compute, with the largest gains on the two zebra subclasses. 
WildBox is released with video-level splits, evaluation code, and baseline checkpoints to enable progress in 3D wildlife perception from drone video.
\end{abstract}

\section{Introduction}
\label{sec:intro}

Drone video-based monitoring is used increasingly in wildlife ecology, conservation, and behavioural research~\cite{tuia_perspectives_2022, wilddrone_frontiers, koger_quantifying_2023, 10.1093/biosci/biaf069, schad_opportunities_2023, wirsing_rapidly_2022, pedrazzi2025advancing}. 
Drones can follow animals over large areas, record dense group interactions, and observe habitats where ground-based methods are incomplete or disruptive to wildlife~\cite{afridi2025impact}. 
While animal behaviour unfolds in three dimensions, computer vision pipelines built on aerial videos remain predominantly two-dimensional: animals are detected~\cite{mou_novel_2024}, counted~\cite{may_polo_2024}, or tracked~\cite{duporge_baboonland_2025} in image space. 
However, such 2D representations remain insufficient for downstream applications that require reasoning about real-world geometry. 
For example, tasks such as viewpoint-aware re-identification~\cite{sundaresan_adapting_2025,grolleau_moo_2026} and drone autonomy~\cite{edou-paams} would benefit from 3D representation of animals.
Parametric mesh models provide rich pose and shape but require taxon-specific templates that do not scale across long-tail taxa. 
3D bounding boxes are species-agnostic, compatible with existing monocular 3D detection architectures, and encode position, extent, orientation, depth ordering, and occlusion structure without requiring full anatomical reconstruction.

Drones used for ecological fieldwork are usually equipped with monocular RGB cameras~\cite{schad_opportunities_2023}. Adding LiDAR, synchronised multi-camera rigs, or calibrated reference objects increases payload, cost, and deployment complexity, limiting their use in typical wildlife monitoring scenarios.
These constraints have driven a line of work aimed at making 3D perception viable from monocular RGB alone, without dedicated 3D sensors or hand-annotated 3D ground truth.
OVM3D-Det~\cite{huangTrainingOpenVocabularyMonocular2024b} trains open-vocabulary monocular 3D detectors from RGB images and automatically generated pseudo-3D labels, motivated by the high deployment cost of LiDAR and other 3D sensors. 
Similarly, LabelAny3D~\cite{yao2025labelany3d} produces pseudo 3D bounding-box annotations for arbitrary categories on in-the-wild images through an analysis-by-synthesis pipeline. 
This is motivated by the limited category coverage of existing 3D datasets and the labour cost of manual 3D annotation. 
Together, these works show that scalable 3D supervision is becoming practical for in-the-wild RGB data, but also confirm that monocular depth estimation and 3D box generation remain the central bottlenecks, especially under domain shift.
Wildlife monitoring data collected with drones presents such strong domain shifts. 
Existing monocular 3D benchmarks such as KITTI~\cite{geigerAreWeReady2012}, nuScenes~\cite{caesarNuScenesMultimodalDataset2020}, Waymo~\cite{Sun_2020_CVPR}, Argoverse~\cite{changArgoverse3DTracking2019}, and Omni3D~\cite{brazilOmni3DLargeBenchmark2023b} are dominated by automotive, indoor, or ground-level imagery. 
Aerial wildlife datasets, by contrast, provide real drone footage but almost exclusively 2D supervision. 
Synthetic drone wildlife datasets and animal mesh datasets are valuable complements, yet none constitutes a benchmark of real monocular drone footage paired with standard-format 3D detection targets. 
This motivates the basic questions whether monocular 3D perception is feasible on real drone-based wildlife video, and how far existing monocular 3D detectors transfer to this domain.

To address these questions, we introduce \textbf{WildBox}, a dataset and benchmark for monocular relative-scale 3D wildlife perception from real monocular drone video.
WildBox comprises six classes of animals native to the African savanna annotated with human-reviewed camera-frame tracked 3D bounding boxes in a per-video segment scale-normalised coordinate frame. 
Annotations preserve relative 3D layout, orientation, extent, projection, and depth ordering within a video segment without asserting metric scale across flights.
To understand whether existing monocular 3D perception transfers to this regime, we evaluate two open-vocabulary architectures with structurally different 3D heads: (a) OVMono3D-LIFT~\cite{yao2024open} (learned Cube R-CNN~\cite{brazilOmni3DLargeBenchmark2023b} with CLIP-text-prompted~\cite{radfordLearningTransferableVisual2021a} classifier) and (b) DetAny3D~\cite{zhang2025detect} (promptable 3D foundation model fusing frozen SAM~\cite{kirillov_segment_2023} and UniDepth~\cite{piccinelli_unidepth_2024}-pretrained DINOv2~\cite{oquab_dinov2_2024} backbones).
We select these baselines because they are the current state-of-the-art exemplars of the two structurally distinct paradigms in open-vocabulary monocular 3D detection, so a shared failure under aerial domain shift cannot be attributed to either paradigm alone.
We test both under two zero-shot 2D-input conditions: open-vocabulary text-prompted detections from GroundingDINO~\cite{liu_grounding_2024}, and ground-truth 2D box prompts as a perfect-detection ceiling. 
Both architectures are then fine-tuned on WildBox; the curriculum and iteration-budget ablations on OVMono3D-LIFT are reported with multi-seed mean\,$\pm$\,std for the two headline configurations.
Open-vocabulary 2D foundation models produce usable wildlife boxes zero-shot (50.55 AP@50), but 3D lifting collapses to 0.00 AP across both architectures and every 2D-input condition tested, including when ground-truth 2D boxes are supplied as prompts; the failure therefore lies in the 3D stage rather than in 2D quality or in a particular 3D-head design. 
Fine-tuning on WildBox recovers measurable 3D performance (13.17 macro AP3D, from 0.00 zero-shot), and a per-axis disentangled error decomposition (image-plane, depth, dimension, pose) attributes 84--99\% of overall normalised Hausdorff distance to depth (over 99\% in zero-shot), isolating monocular aerial depth as the dominant failure mode. 
We further find that, when video segment numbers are asymmetric across taxonomically related categories, pretraining on a merged superclass before fine-tuning on subclass labels improves macro 3D performance with less total compute, with the largest gains on the merged-then-split sibling subclasses.
Absolute metric lies outside the scope of this benchmark without additional scale supervision.

\section{Related Work}
\label{sec:sota}

\paragraph{Monocular 3D object detection.}
Closed-vocabulary monocular 3D detection has matured around automotive benchmarks~\cite{geigerAreWeReady2012,caesarNuScenesMultimodalDataset2020,Sun_2020_CVPR,changArgoverse3DTracking2019} and has been extended to indoor and large-scale unified training~\cite{brazilOmni3DLargeBenchmark2023b}. 
Open-vocabulary extensions lift 2D detections from text-prompted detectors~\cite{liu_grounding_2024}
into 3D via learned lifters~\cite{yao2024open} or via promptable foundation models~\cite{zhang2025detect}. 
However, all of these approaches inherit the perspective distribution of their training data, and their transfer to aerial imagery has not yet been systematically evaluated.
LabelAny3D~\cite{yao2025labelany3d} introduced both a pseudo-labelling pipeline for in-the-wild 3D annotation and a scale-invariant evaluation protocol (Rel-AP3D) for the resulting pseudo-labels. 
We adopt Rel-AP3D directly, however its labelling pipeline is not feasible at WildBox's frame scale; we document the comparison in Appendix~\ref{app:annotation_scaling}.
Pseudo-3D supervision derived from open-vocabulary 2D models and monocular depth has been shown to scale~\cite{huangTrainingOpenVocabularyMonocular2024b}, but inherits the noise of ground-level monocular depth estimators under domain shift. 
WildBox uses feed-forward 3D reconstruction~\cite{shuklaWildLIFTLiftingMonocular2026} instead, which is better suited to aerial perspective.

\paragraph{Aerial wildlife datasets.}
Existing drone wildlife datasets provide 2D annotations across a range of tasks: thermal multi-object tracking~\cite{bondiBIRDSAIDatasetDetection2020}, aerial
censuses~\cite{naudeAerialElephantDataset2019}, species
detection~\cite{mouWAIDLargeScaleDataset2023a}, behavioural recognition on Kenyan
ungulates~\cite{skovorodnikovFERALVideoUnderstandingSystem2025, kholiavchenko_kabr_2024}, multi-UAV blackbuck
tracking~\cite{naikBuckTalesMultiUAVDataset2024}, onboard real-time
tracking~\cite{datWildLiveRealtimeVisual2025}, and multi-site low-altitude
video~\cite{klineMMLAMultiEnvironmentMultiSpecies2025}. 
None of these datasets provides 3D supervision.
KABR~\cite{kholiavchenko_kabr_2024, KABR_Raw_Videos, kline2025kabrtoolsautomatedframeworkmultispecies} is the closest in terms of domain (Kenyan ungulates observed in-situ from drones) and establishes the convention of separating plains and Grévy's zebra, which we follow.
We extend the set of classes to include giraffes, rhinos, elephants, and gazelles.

\paragraph{3D supervision for animals.}
Besides research based on aerial drone data, other works consider reconstructing (single) animals using 3D models. 
To achieve reconstruction, animal 3D annotation is crucial and is usually divided into three paradigms: SMAL-based mesh fitting on web and staged imagery~\cite{xu_animal3d_2023,zuffiThreeDSafariLearning2019a}, multi-view keypoint rigs in laboratory and controlled enclosures~\cite{dunnGeometricDeepLearning2021,marshall_pair-r24m_2021,naik3DPOPAutomatedAnnotation2023,9561338}, and LiDAR-assisted ground capture~\cite{muramatsu_wildpose_2024,aamirWildDepthMultimodalDataset2026}. 
None of these works operate on monocular drone video. 
Synthetic drone-rendered 3D wildlife imagery is available~\cite{bonetto_synthetic_2023}, but has not been paired with a real-world counterpart.
\citet{shukla_towards_2024} demonstrate SMAL fitting on oblique drone footage of plains zebras without releasing a dataset.

\paragraph{Position of WildBox.}
Table~\ref{tab:dataset_positioning} places WildBox at the intersection of the three literatures. 
To our knowledge, no prior dataset combines real-world drone imagery, multiple species of wildlife, and KITTI-style 3D bounding boxes. 

\begin{table}[t]
\centering
\caption{Positioning of WildBox relative to aerial wildlife datasets, animal 3D datasets, and monocular 3D detection benchmarks.}
\label{tab:dataset_positioning}
\footnotesize
\setlength{\tabcolsep}{3.8pt}
\renewcommand{\arraystretch}{1.06}
\begin{tabular}{@{}lccc lcr@{}}
\toprule
 & \multicolumn{3}{c}{Domain} & \multicolumn{3}{c}{3D supervision} \\
\cmidrule(lr){2-4} \cmidrule(l){5-7}
Dataset & Real & Aerial & Wildlife & Source & Std. 3D box & \# 3D inst. \\
\midrule
\multicolumn{7}{@{}l}{\emph{Aerial wildlife datasets}} \\
BIRDSAI~\cite{bondiBIRDSAIDatasetDetection2020}       & $\sim$ & \cmark & \cmark & --          & \xmark & -- \\
Aerial Elephant~\cite{naudeAerialElephantDataset2019} & \cmark & \cmark & \cmark & --          & \xmark & -- \\
WAID~\cite{mouWAIDLargeScaleDataset2023a}             & \cmark & \cmark & \cmark & --          & \xmark & -- \\
KABR~\cite{kholiavchenko_kabr_2024}                   & \cmark & \cmark & \cmark & --          & \xmark & -- \\
BuckTales~\cite{naikBuckTalesMultiUAVDataset2024}     & \cmark & \cmark & \cmark & --          & \xmark & -- \\
WildLive~\cite{datWildLiveRealtimeVisual2025}         & \cmark & \cmark & \cmark & --          & \xmark & -- \\
MMLA~\cite{klineMMLAMultiEnvironmentMultiSpecies2025} & \cmark & \cmark & \cmark & --          & \xmark & -- \\
\addlinespace[2pt]

\multicolumn{7}{@{}l}{\emph{Animal 3D datasets}} \\
Animal3D~\cite{xu_animal3d_2023}        & \cmark & \xmark & \cmark & Mesh/Kpts    & \xmark & 3379 \\
WildPose~\cite{muramatsu_wildpose_2024} & \cmark & \xmark & \cmark & LiDAR        & \xmark & -- \\
WildDepth~\cite{aamirWildDepthMultimodalDataset2026}
                                         & \cmark & \xmark & \cmark & LiDAR        & \xmark & -- \\
\addlinespace[2pt]

\multicolumn{7}{@{}l}{\emph{Synthetic aerial wildlife 3D}} \\
GRADE~\cite{bonetto_grade_2026} & \xmark & \cmark & \cmark & Synthetic    & $\sim$ & -- \\
\addlinespace[2pt]

\multicolumn{7}{@{}l}{\emph{Monocular 3D detection benchmarks}} \\
KITTI~\cite{geigerAreWeReady2012}       & \cmark & \xmark & \xmark & LiDAR        & \cmark & $\sim$80k \\
Omni3D~\cite{brazilOmni3DLargeBenchmark2023b}
                                         & \cmark & \xmark & \xmark & LiDAR/RGB-D  & \cmark & $\sim$3M$^{\ddagger}$ \\
LabelAny3D/COCO3D~\cite{yao2025labelany3d}
                                         & \cmark & \xmark & $\sim$ & Model+human  & \cmark & 5.4k \\
\midrule
\textbf{WildBox (ours)}                  & \cmark & \cmark & \cmark & Model+human  & \cmark & \textbf{237.5k} \\
\bottomrule
\end{tabular}

\vspace{2pt}
\begin{minipage}{0.96\linewidth}
\scriptsize
\emph{Notes.} ``Std. 3D box'' denotes standard-format 3D bounding boxes compatible with monocular 3D detection evaluation. 
Model+human denotes model-in-the-loop 3D box proposal with human review.
$\sim$ denotes partial satisfaction: BIRDSAI mixes real and synthetic footage, GRADE releases 3D boxes in a non-benchmark format, and COCO3D contains some animal categories without being wildlife-focused.
$^{\ddagger}$Omni3D aggregates KITTI~\cite{geigerAreWeReady2012}, nuScenes~\cite{caesarNuScenesMultimodalDataset2020}, SUN RGB-D~\cite{song_sun_2015}, ARKitScenes~\cite{arkitscenes}, Hypersim~\cite{hypersim}, and Objectron~\cite{ahmadyan_objectron_2021}; none is aerial or wildlife-focused.
\end{minipage}
\end{table}

\section{The WildBox Dataset}
\label{sec:dataset}

WildBox aggregates oblique drone footage acquired with DJI drones during field campaigns conducted between 2023 and 2026 at Ol Pejeta Conservancy in Kenya, and at Bristol Zoological Society in the UK. 
All data was collected following applicable local laws and under valid permits, and efforts were made to minimize disturbance to wildlife.
Videos from KABR dataset~\cite{KABR_Raw_Videos} collected at Mpala Research Centre are also used.
WildBox annotates seven species grouped into six benchmark classes: reticulated giraffe, plains zebra, Grévy's zebra, African bush elephant, Thomson's gazelle, and rhinoceros. 
The rhino class combines black and white rhinoceros.
The binary species attribute is not used by the present benchmark and both black and white rhino instances were assigned a single label “rhino”. 
Scientific names and IUCN conservation status for each species are listed in Appendix~\ref{app:species}.
The dataset comprises 64 videos (51 training, 13 validation) divided into 345 segments of up to 200 frames each (263 training, 82 validation).
The frames are sampled at 10 Hz, with a total of 59{,}758 frames, and 237{,}505 3D bounding box instances (Figure~\ref{fig:data_overview}). 
Splits are constructed at the video level with stratification by species, preventing within-sequence leakage of background, herd, or animal identity.

\begin{figure}[t]
\centering
\includegraphics[width=\textwidth]{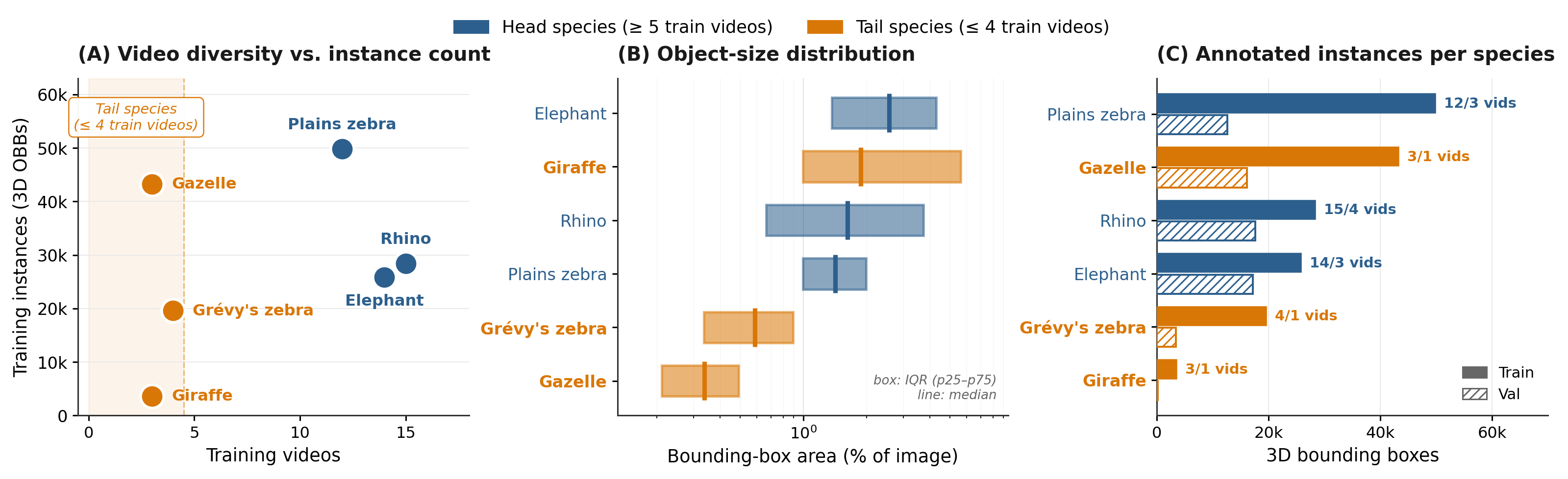}
\caption{WildBox dataset characterisation. 
(A) Per-class training-video count vs.\ training-instance count: tail classes (3--4 train videos) include both classes with sparse instances (giraffe) and classes with high instance density (gazelle). OBBs refers to oriented bounding boxes.
(B) Bounding-box area per class (\% of image, log-scale, IQR with median bar): aerial perspective produces two orders of magnitude spread in object size across taxa. 
(C) Annotated instance counts per class, train (solid) and validation (hatched), with video counts annotated. Tail classes are highlighted in orange. 
Giraffe val (110 instances) is below visual resolution; see Table~\ref{tab:full_inventory} in Supplementary for exact counts.}
\label{fig:dataset_overview}
\end{figure}

The class distribution is not artificially rebalanced.
Figure~\ref{fig:dataset_overview} characterises the dataset along three axes: per-species video diversity vs.\ instance count (Panel~A), bounding-box areas distribution per class (Panel~B), and total annotated instances with train/validation breakdown (Panel~C). 
Two structural features matter for the benchmark. 
The major classes (plains zebra, rhino, elephant) have 12--15 training videos each, while rarer classes (giraffe, Grévy's zebra, gazelle) have only 3--4; we do not rebalance, since a deployed drone-monitoring system sees the distribution its flights produce. 
Crucially, instance-count and video-count are decoupled: gazelle has 43{,}280 training instances across just 3 training videos.
Gazelle class has more instances than rhinos (28{,}381, 15 videos), but a fifth of the video-level diversity.
The per-class evaluation protocol (see Section \ref{sec:eval}) reports performance under both counts, and we show empirically (Section \ref{sec:experiments}) that video-level diversity appears more predictive of 3D detection performance than raw instance counts.

\begin{figure}
    \centering
    \includegraphics[width=0.75\linewidth]{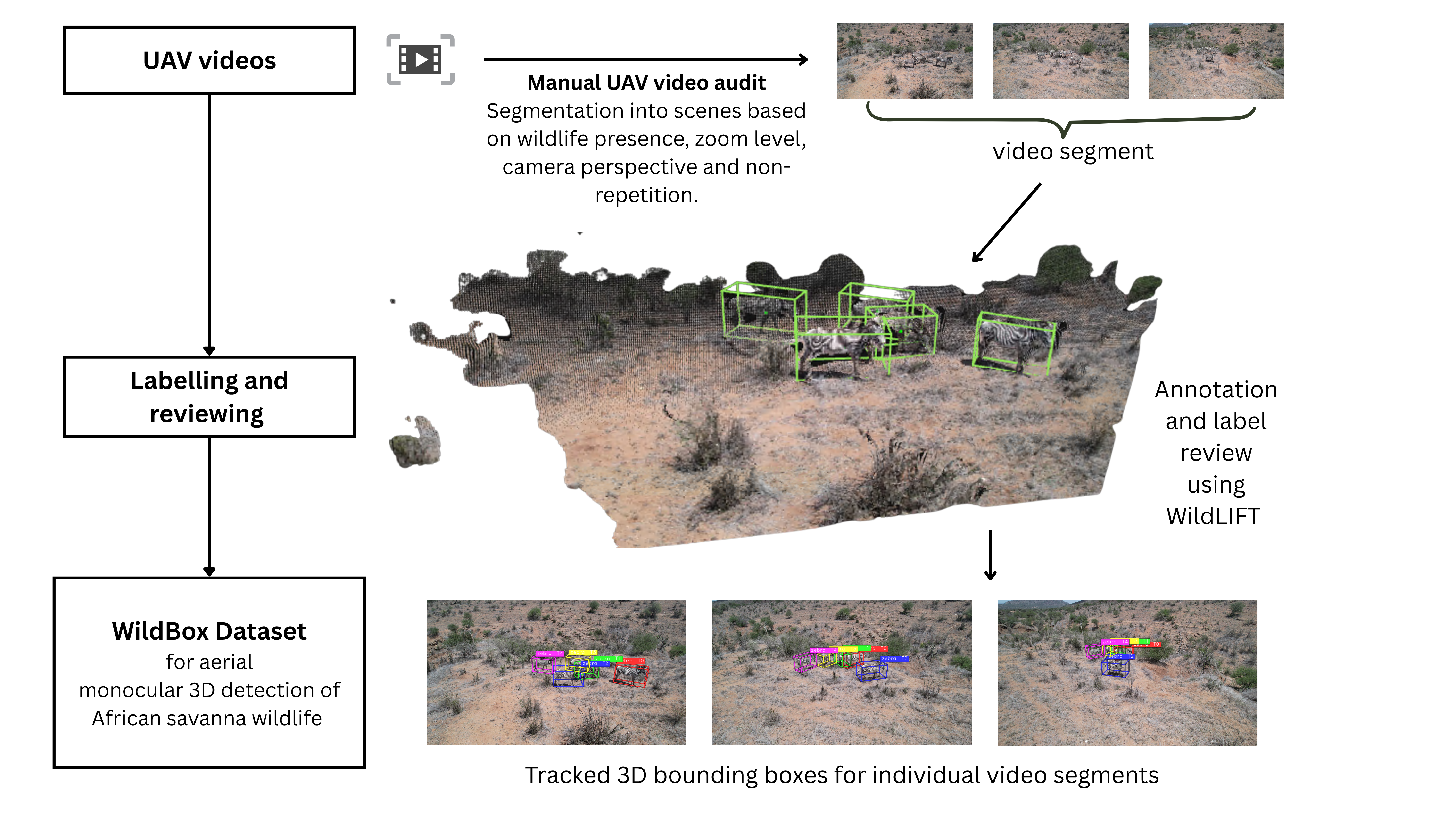}
    \caption{WildBox annotation pipeline. 
    Raw drone footage~\cite{KABR_Raw_Videos, kline2025kabrtoolsautomatedframeworkmultispecies} is first manually audited and segmented into clips containing wildlife under varied oblique viewpoints, with redundant or repetitive sequences discarded. 
    Each segment is then processed with WildLIFT~\cite{shuklaWildLIFTLiftingMonocular2026}: WildLIFT-RT reconstructs the scene as a dense 3D pointmap and produces identity-consistent 3D tracks, while WildLIFT-A fits oriented 3D bounding boxes to each tracked individual. 
    All boxes are then reviewed and refined frame-by-frame in the WildLIFT-A interface to yield the final tracked 3D annotations that constitute the WildBox dataset.}
    \label{fig:anno_method}
\end{figure}

\paragraph{Video segments and tracks.}
Each raw video is first manually audited and divided into shorter clips called \emph{segments}, with boundaries chosen to keep wildlife presence, viewpoint, and zoom level stable within a clip and to discard redundant or repetitive content (Figure~\ref{fig:anno_method}, top). 
Segments are capped at 200 frames; the 51/13 video split decomposes into 263/82 segments (Table~\ref{tab:full_inventory}). 
A track is one animal instance followed through one segment, keyed (video, segment, track\_id); an annotation is one animal in one frame. 
The total number of tracks was 1478 and a track spans 160.7 frames on average (median 200).
The same physical animal may be re-tracked in a later segment, as identities are not maintained across segments.
Further track level details are available in Appendix \ref{app:dataset}.
The segment is the unit of 3D reconstruction (one dense pointmap per segment), of tracking, and of the per-segment scale normalisation described below.

\paragraph{Annotation method.}

WildBox uses a \emph{model-in-the-loop} annotation pattern (Figure~\ref{fig:anno_method}): WildLIFT~\cite{shuklaWildLIFTLiftingMonocular2026} proposes 3D oriented bounding boxes which human reviewers refine and confirm frame-by-frame; only human-confirmed boxes enter the release (more details in Appendix \ref{app:human_review} and labelling visualisation in Figure \ref{fig:geometry_label} in \ref{app:anno_vis}). 
The pattern is standard for in-the-wild monocular 3D annotation where exhaustive manual 3D bounding box fitting is impractical~\cite{yao2025labelany3d}. Appendix~\ref{app:annotation_scaling} compares WildLIFT to per-image alternatives.

\paragraph{Annotation target.}
WildBox provides tracked 3D bounding box labels. 
Each box is parameterised by camera-frame position, extent, and a full $3{\times}3$ rotation, and carries an instance ID that is consistent across frames of the same segment. 
Cross-frame identity supports identity-aware downstream tasks but is not benchmarked in this paper.

\paragraph{Coordinate system and scale regime.}
All 3D annotations are expressed in the camera frame (X right, Y down, Z forward) and follow the Omni3D ordering convention for dimensions.
Rotations are stored as full $3 \times 3$ matrices, since drone-oblique footage carries substantial orientation variation in pitch and roll that yaw-only parameterisations cannot capture. 
Per-segment scale is normalised by dividing all 3D coordinates by the median absolute camera-space depth of the segment's reconstruction; this preserves box shape, orientation, 3D IoU, and 2D projection within a segment, but renders absolute depth incomparable across segments. 
Detailed per-class inventory tables and the validation-split sensitivity analysis are provided in Appendix~\ref{app:dataset}.

\section{Evaluation}
\label{sec:eval}

We evaluate two open-vocabulary monocular 3D detectors with structurally different 3D heads: \textbf{OVMono3D-LIFT}~\cite{yao2024open}, a Cube R-CNN cube head with a CLIP-text-prompted classifier, and \textbf{DetAny3D}~\cite{zhang2025detect}, a promptable foundation-model 3D interpreter built on SAM-ViT-H and DINOv2 ViT-L/14.

\paragraph{Settings.}
We evaluate the detectors under four settings.
\textbf{(i) Closed-vocabulary zero-shot} uses the architecture's own 2D stage trained on Omni3D~\cite{brazilOmni3DLargeBenchmark2023b}; classifier outputs do not overlap with wildlife categories and are filtered out at evaluation.
This setting tests whether the pretrained 2D stage generalises to wildlife at all. 
Available for OVMono3D-LIFT only; DetAny3D's text-conditioned 2D stage has no closed-vocabulary equivalent.
\textbf{(ii) Open-vocabulary zero-shot} replaces the architecture's 2D proposals with pre-computed text-prompted detections from GroundingDINO~\cite{liu_grounding_2024}. 
This is the OVMono3D zero-shot protocol of~\cite{yao2024open} and the analogous DetAny3D inference path. 
It assesses whether the 3D head transfers when 2D localisation is supplied by a foundation model.
\textbf{(iii) Ground-truth 2D box prompt zero-shot} feeds the 3D head with the WildBox ground-truth 2D boxes and class labels. 
Since the 2D inputs are perfect by construction, this setting isolates 3D-stage error and serves as a ceiling on zero-shot 3D lifting.
\textbf{(iv) Supervised fine-tuning} initialises from the same Omni3D-pretrained checkpoint and trains on the WildBox training split.
Settings (ii)--(iv) are available for both architectures.

\paragraph{Training configurations.}
Within the supervised fine-tuning setting we compare two initialisation strategies for OVMono3D-LIFT. 
\emph{Direct fine-tuning} trains the full six-class label scheme from the Omni3D-pretrained checkpoint.
\emph{Curriculum-initialised fine-tuning} first pretrains on a five-class label scheme that merges plains zebra and Grévy's zebra into a single ``zebra'' superclass, then continues training on the six-class split labels. 
The two strategies are compared with the curriculum-init using 5k fewer total training iterations than direct fine-tuning at its longest schedule (Table~\ref{tab:ablations}). 
DetAny3D is fine-tuned for two epochs; performance at one and two epochs is comparable, while three or more epochs overfits (Appendix~\ref{app:detany3d_epochs}). 
Throughout, REPEAT\_THRESHOLD ($t_R$) controls the RepeatFactorTrainingSampler used for class-balanced sampling: rare classes with image-level frequency below $t_R$ are oversampled by a factor proportional to $\sqrt{t_R / f_c}$. 
The five-class merged-zebra checkpoint is trained on the same WildBox training data with the plains and Gr\'evy's zebra labels merged; the data is unchanged, only the label space.

\paragraph{Metrics.}
\textbf{AP-BEV} at IoU thresholds 0.25 and 0.50 is our primary 3D metric:
2D AP on rotated-rectangle bird's-eye-view footprints obtained by dropping the camera-Y axis. 
AP-BEV is robust to the per-segment depth-axis scale normalisation in the ground truth. 
\textbf{AP3D} at standard thresholds is reported for continuity with the monocular 3D literature, but is secondary because cross-segment absolute-depth comparisons are unstable under our scale regime.
\textbf{Rel-AP3D}~\cite{yao2025labelany3d} is reported with a global scale-alignment grid search on $s \in [0.05, 3.0]$ at 32 points.
\textbf{2D AP@50} is reported (COCO AP at IoU $=0.5$) to isolate 2D localisation from 3D lifting. 
The two architectures use different 2D-AP conventions: OVMono3D-LIFT scores the projected 3D bounding boxes, so under GT-2D box prompt zero-shot the value reflects cube-head degradation rather than the perfect 2D input. 
DetAny3D scores the input prompt box, so its 2D AP is unchanged by 3D-head fine-tuning. 
The disentangled \textbf{normalised Hausdorff distance (NHD)} is reported as a diagnostic, decomposed by leave-one-in into image-plane (xy), depth (z), dimension, and pose components, each pair-normalised by the GT cuboid diagonal.

\paragraph{Reporting.}
We report per-class AP for every metric, alongside micro- and macro-averages. 
Macro-average is the primary summary statistic because methods that gain head-class AP at the cost of rare-class AP show a decrease in macro AP. 
For the two headline OVMono3D-LIFT fine-tuned configurations (direct at 15k iterations and curriculum-init at 10k continuation), we report mean and standard deviation across three seeds. 
Remaining ablation configurations are single-seed and serve to confirm that variance from sampling and iteration-budget choices is small relative to the gaps between conditions.

\section{Results}
\label{sec:experiments}

\begin{table}[t]
\centering
\caption{Main benchmark on WildBox validation. 
Bold marks the recommended fine-tuned configuration.}
\label{tab:main_benchmark}
\footnotesize
\setlength{\tabcolsep}{3.5pt}
\renewcommand{\arraystretch}{1.08}
\newcommand{\mstd}[2]{#1\,{\scriptsize$\pm$\,#2}}
\begin{tabular}{@{}llccccc@{}}
\toprule
Model & Setting & 2D AP@50 & AP-BEV@0.25 & AP-BEV@0.50 & AP3D & Rel-AP3D \\
\midrule
\multicolumn{7}{@{}l}{\emph{Zero-shot}} \\
LIFT     & Closed-vocab        &  1.66 & 0.00 & 0.00 & 0.00 & $\dagger$ \\
LIFT     & OV, GDino           & 50.55 & 0.00 & 0.00 & 0.00 & 0.01 \\
LIFT     & GT-2D prompt        & 72.28 & 0.01 & 0.00 & 0.00 & 0.10 \\
DetAny3D & OV, GDino           & 60.89 & 0.00 & 0.00 & 0.00 &  0.01 \\
DetAny3D & GT-2D prompt        & 94.11 & 0.00 & 0.00 & 0.00 & 0.06 \\
\midrule
\multicolumn{7}{@{}l}{\emph{Supervised fine-tuning}} \\
LIFT     & Direct              &
  \mstd{77.77}{0.87} &
  \mstd{19.61}{1.69} &
  \mstd{4.73}{1.80} &
  \mstd{9.12}{0.84} &
  \mstd{8.88}{0.89} \\
\textbf{LIFT} & \textbf{Curriculum-init} &
  \textbf{\mstd{80.41}{0.37}} &
  \textbf{\mstd{25.74}{1.41}} &
  \textbf{\mstd{8.68}{0.47}} &
  \textbf{\mstd{13.17}{0.69}} &
  \textbf{\mstd{12.89}{0.68}} \\
DetAny3D & 2 epochs &
  60.89 & 8.33 & 1.99 & 4.15 & 4.15 \\
\bottomrule
\end{tabular}

\vspace{2pt}
\begin{minipage}{0.96\linewidth}
\scriptsize
\emph{Notes.} LIFT = OVMono3D-LIFT; OV = open-vocabulary; GDino = GroundingDINO.
All 3D metrics are macro-averaged. LIFT fine-tuned rows report 3-seed mean$\pm$std.
Rel-AP3D applies global scale alignment over $s \in [0.05,3.0]$.
$^{\dagger}$Rel-AP3D is undefined because closed-vocabulary predictions do not yield a usable matched-pair set.
\end{minipage}
\end{table}

\begin{table}[t]
\centering
\caption{Disentangled NHD decomposition on WildBox validation. Lower is better.}
\label{tab:nhd}
\footnotesize
\setlength{\tabcolsep}{4.2pt}
\renewcommand{\arraystretch}{1.08}
\begin{tabular}{@{}llrrrrrr@{}}
\toprule
Model & Setting & $xy$ & $z$ & dim. & pose & overall & $z$/overall \\
\midrule
\multicolumn{8}{@{}l}{\emph{Zero-shot}} \\
LIFT     & Closed-vocab     &  12.25 &  143.34 &  15.41 & 1.21 &  144.78 & 99.0\% \\
LIFT     & OV, GDino        &  45.65 &  561.40 &  52.01 & 1.46 &  565.69 & 99.2\% \\
LIFT     & GT-2D prompt     &  50.50 &  625.27 &  49.55 & 1.47 &  629.70 & 99.3\% \\
DetAny3D & OV, GDino        & 201.70 & 3021.61 & 182.61 & 1.44 & 3035.11 & 99.6\% \\
DetAny3D & GT-2D prompt     & 181.12 & 2635.78 & 158.48 & 1.45 & 2648.48 & 99.5\% \\
\midrule
\multicolumn{8}{@{}l}{\emph{Supervised fine-tuning}} \\
LIFT     & Direct           &   2.14 &    6.43 &   0.86 & 0.58 &    7.52 & 85.5\% \\
\textbf{LIFT} & \textbf{Curriculum-init}
                             & 2.19 & \textbf{5.97} & \textbf{0.76} & \textbf{0.56} & \textbf{7.05} & \textbf{84.7\%} \\
DetAny3D & 2 epochs & \textbf{1.75} &   11.48 &   1.01 & 1.79 &   12.35 & 92.9\% \\
\bottomrule
\end{tabular}

\vspace{2pt}
\begin{minipage}{0.96\linewidth}
\scriptsize
\emph{Notes.} LIFT = OVMono3D-LIFT; OV = open-vocabulary; GDino = GroundingDINO.
``Overall'' is the full corner-Hausdorff distance per matched pair.
Component columns are leave-one-in NHDs; $z$/overall is the depth share of total NHD.
LIFT fine-tuned rows report 3-seed means. 
DetAny3D fine-tuning is reported from a single seed.
\end{minipage}
\end{table}

\begin{table}[t]
\centering
\caption{OVMono3D-LIFT ablations on WildBox validation. Configuration D is the recommended fine-tuned setting.}
\label{tab:ablations}
\footnotesize
\setlength{\tabcolsep}{4.5pt}
\renewcommand{\arraystretch}{1.08}
\newcommand{\mstd}[2]{#1\,{\scriptsize$\pm$\,#2}}
\begin{tabular}{@{}clcccccc@{}}
\toprule
 &  & \multicolumn{4}{c}{Training} & \multicolumn{2}{c}{Macro AP} \\
\cmidrule(lr){3-6}\cmidrule(l){7-8}
ID & Configuration & Init & $t_R$ & 6-cls iters & Total iters & AP-BEV@0.50 & AP3D \\
\midrule
A & Direct, baseline  & Omni3D & 0.50 & 15k & 15k &
  \mstd{4.73}{1.80} & \mstd{9.12}{0.84} \\
B & Direct, lower $t_R$  & Omni3D & 0.35 & 15k & 15k &
  5.25 & 8.32 \\
C & Direct, extended  & Omni3D & 0.50 & 25k & 25k &
  4.17 & 10.76 \\
D & \textbf{Curriculum-init} & 5-cl ckpt & 0.50 & 10k & 20k &
  \textbf{\mstd{8.68}{0.47}} & \textbf{\mstd{13.17}{0.69}} \\
\bottomrule
\end{tabular}

\vspace{2pt}
\begin{minipage}{0.94\linewidth}
\scriptsize
\emph{Notes.} $t_R$ is the RepeatFactorTrainingSampler threshold. 
A and D report 3-seed mean$\pm$std; B and C are seed 0.
D uses a 10k-iteration five-class merged-zebra checkpoint, followed by 10k iterations of six-class continuation, for 20k total iterations.
\end{minipage}
\end{table}

Table~\ref{tab:main_benchmark} reports headline results for both architectures under all settings. 
We unpack the table in three steps:
the architecture-independence of the zero-shot collapse (\S\ref{sec:exp:headline}), the recovery under fine-tuning (\S\ref{sec:exp:ft}), and the disentangled error decomposition (\S\ref{sec:exp:depth}). 
We then report per-class disaggregation (\S\ref{sec:exp:class}) and a curriculum-initialisation ablation (\S\ref{sec:exp:curriculum}). 
We also report how performance varies with acquisition conditions in Appendix~\ref{app:per_acqui}, and the micro-level results in Appendix~\ref{app:micro}.
Full training hyperparameters and compute details are provided in Appendix~\ref{app:repro}.

\subsection{Zero-shot 3D lifting collapses across architectures and 2D sources}
\label{sec:exp:headline}

Open-vocabulary 2D foundation models recover wildlife localisation, but zero-shot 3D lifting collapses regardless of 2D source or 3D-head architecture.
With open-vocabulary 2D inputs, OVMono3D-LIFT achieves 50.55 2D AP@50, indicating that text-prompted 2D foundation models yield usable animal boxes in this domain. 
The corresponding zero-shot 3D AP is 0.00. 
The same collapse holds when the 2D stage is replaced with the closed-vocabulary Omni3D-pretrained RPN (1.66 2D AP@50, 0.00 AP3D) and when ground-truth 2D boxes are supplied as prompts (72.28 2D AP@50, 0.00 AP3D). 
The failure cannot be attributed to closed-vocabulary classification, GroundingDINO proposal noise, or 2D-detection quality.
Qualitatively, the zero-shot 3D head produces inflated, near-uniform cuboids that ignore class scale and inter-instance depth ordering (Figure~\ref{fig:qual}, Appendix~\ref{app:qualitative}).

The same pattern transfers to DetAny3D, a structurally distinct 3D detector. 
Open-vocabulary zero-shot yields 60.89 macro 2D AP@50 and 0.00 AP3D.
Under ground-truth 2D box prompts the 2D inputs reach 94.11 macro 2D AP@50 by construction, but AP3D remains 0.00. 
The two architectures use structurally different 3D heads --- a Cube R-CNN cube head versus a promptable foundation-model 3D interpreter --- so the collapse is not specific to a single 3D-head design. 
The remaining shared factor is the pretraining-distribution mismatch between Omni3D's ground-level/indoor imagery and aerial wildlife geometry.
The Rel-AP3D column further constrains the failure: 
under optimal global scale alignment over $s \in [0.05, 3.0]$, both architectures remain below 0.1 macro Rel-AP3D in zero-shot. 
Uniform rescaling does not recover 3D performance --- per-prediction shape and orientation errors prevent meaningful 3D-IoU even at the best global scale.

\subsection{Fine-tuning on WildBox recovers measurable 3D performance}
\label{sec:exp:ft}

Supervised fine-tuning on WildBox lifts both architectures from 0.00 to non-zero 3D AP, with curriculum-initialised training giving the strongest result on OVMono3D-LIFT.
OVMono3D-LIFT recovers to 8.68\,$\pm$\,0.47 AP-BEV@0.50 macro and 13.17\,$\pm$\,0.69 AP3D macro under the curriculum-initialised configuration. 
Direct fine-tuning from the Omni3D checkpoint at 15k iterations reaches 4.73\,$\pm$\,1.80 AP-BEV@0.50 macro and 9.12\,$\pm$\,0.84 AP3D macro.
DetAny3D, fine-tuned for two epochs, reaches 1.99 AP-BEV@0.50 macro
and 4.15 AP3D macro.

\subsection{Depth dominates the residual error budget}
\label{sec:exp:depth}
Depth (z) accounts for 84--99\% of overall NHD across every condition we evaluate, including those with ground-truth 2D inputs.
Table~\ref{tab:nhd} reports the disentangled NHD decomposition. 
Across zero-shot conditions, depth contributes 99.0--99.6\% of overall NHD; under fine-tuned conditions, depth still contributes 84.7--92.9\%. 
Even under DetAny3D with ground-truth 2D box prompts, where 2D localisation is perfect, NHD-z reaches 2{,}635 --- two orders of magnitude larger than the fine-tuned NHD-z values of 5.97--11.48. 
Perfect 2D supervision does not on its own correct monocular aerial depth: the depth axis carries the bulk of the 3D error budget even when every other input is correct. 
Our benchmark therefore isolates aerial monocular depth estimation as the open problem that limits 3D wildlife perception in this regime.

\subsection{Per-class results suggest video diversity matters more than instance count}
\label{sec:exp:class}

Per-class fine-tuned 3D AP shows a stronger association with training-video count than with training-instance count.
Table~\ref{tab:per_class_ft} (Appendix~\ref{app:per_class_ft}) reports per-class fine-tuned performance under the curriculum-initialised configuration (3-seed mean\,$\pm$\,std).
Plains zebra (12 training videos) reaches 33.40\,$\pm$\,0.34 AP3D. 
Gazelle has 3 training videos but 43{,}280 training instances, more than $1.5\times$ rhino's 28{,}381 instances spread across 15 videos; gazelle nevertheless reaches only 2.49\,$\pm$\,0.22 AP3D, below rhino's 5.43\,$\pm$\,0.29. 
The comparison isolates video-level diversity from instance count and we interpret the result as evidence that variation across recording conditions (lighting, terrain, viewing angle, herd composition) is more predictive of 3D generalisation than repeated labels from a small number of videos. 
We report this as a dataset-design implication, not a controlled causal claim.

\subsection{Curriculum initialisation improves performance with less total compute}
\label{sec:exp:curriculum}
Curriculum-initialised fine-tuning outperforms direct fine-tuning with less total compute, and gains concentrate on the merged-then-split rare subclass. 
Table~\ref{tab:ablations} compares four OVMono3D-LIFT training configurations. 
Curriculum-initialised fine-tuning (D) reaches 13.17\,$\pm$\,0.69 macro AP3D and 8.68\,$\pm$\,0.47 macro AP-BEV@0.50 with 10k iterations of 6-class continuation from a 10k-iteration 5-class pretrained checkpoint (20k total). 
Direct fine-tuning from the Omni3D checkpoint reaches 9.12\,$\pm$\,0.84 macro AP3D at 15k (A) and 10.76 at 25k (C); the curriculum exceeds the longer direct run by +2.4 AP3D and +4.5 AP-BEV@0.50 with 5k fewer total iterations.

The gain is non-uniform across classes (per-class breakdown in Appendix~\ref{app:curriculum_per_class}).
Gr\'evy's zebra gains +9.5 AP3D over direct (A). It is the rarer of the two zebra subclasses; under curriculum initialisation, the merged
``zebra'' superclass aggregates training data from both subclasses before they are split. 
Plains zebra, the head class within the merged superclass, gains +9.8 AP3D. 
Singleton classes that were not merged (elephant, rhino) gain 1--2 AP3D. 
Gazelle improves by +1.0 AP3D. 
Giraffe shows variance dominated by single-video val-set difficulty rather than a clear curriculum effect.

Extra direct iterations do not reproduce the curriculum's gain.
Configuration C (25k direct) raises macro AP3D from 9.12 to 10.76 over A (15k direct), but the per-class rank ordering is identical to A (Spearman $\rho = 1.00$ between A and C, vs.\ $\rho = 0.83$ between either and D), and the AP3D gain over A concentrates on giraffe
(+9.7); the remaining three non-zebra classes are within $\pm 0.7$ AP3D of A. 
On AP-BEV@0.50, extra direct iterations are actively harmful: C reaches 4.17 macro, below A's 4.73. 
Curriculum init (D) instead reshapes the class profile and exceeds C on AP-BEV@0.50 across
every class. 
We attribute the improvement to shape priors learned from the merged superclass and retained when the labels are split. 
This single merged-pair experiment (zebras) does not establish that the curriculum benefit generalises; we report it as evidence for the zebra case and as a candidate strategy worth testing on other taxonomically similar pairs in drone-video datasets.

\section{Limitations}
\label{sec:limitations}
WildBox has several limitations. First, the present benchmark evaluates monocular 3D detection frame by frame. 
It does not use temporal cues from video segments, although such cues may improve depth, orientation, and scale consistency. 
Second, WildBox provides instance identities within each segment, but the benchmark does not evaluate 3D tracking or identity-aware prediction. 
These annotations are released to support future work and research in video-based 3D perception, re-identification, and trajectory modelling.
Third, the class taxonomy is not uniform in taxonomic granularity.
Plains zebra and Grévy's zebra are evaluated as separate benchmark classes, following prior aerial wildlife work and because both are represented in the dataset. 
The rhino benchmark class instead merges black and white rhinoceros, although the annotations retain a binary species attribute. 
As a result, the current benchmark tests one subclass split in detail, but does not establish how well the proposed curriculum strategy generalises to other taxonomically related pairs.
Finally, WildBox uses per-segment relative-scale 3D annotations. 
This matches the practical constraints of monocular drone footage, but it does not provide absolute metric scale across flights. 
Methods that require globally metric depth or cross-segment metric consistency will therefore need additional scale supervision.

\section{Conclusion}
\label{sec:conclusion}

We introduced WildBox, a dataset and benchmark for aerial monocular 3D detection of African savanna wildlife from real drone video. 
To our knowledge, WildBox is the first dataset to combine real drone footage, multiple wildlife species, and tracked 3D bounding box labels on video segments (Table~\ref{tab:dataset_positioning}). 
It provides 237{,}505 human-reviewed labels with segment-level instance identities in video-level splits.
Our experiments show that open-vocabulary 2D models localise wildlife in drone imagery, but zero-shot monocular 3D lifting fails across two architectures and across all tested 2D-input
conditions. 
Fine-tuning on WildBox recovers non-zero 3D performance, while the error decomposition identifies aerial monocular depth as the dominant remaining failure mode.
WildBox is intended to support research beyond the frame-wise baselines reported here. 
The dataset enables work on monocular 3D detection of wildlife, tracked 3D wildlife perception, viewpoint-aware re-identification, and wildlife-aware drone autonomy. 
The selected species and environmental conditions represent an important and practically relevant conservation-monitoring scenario. 
Grassland habitats and their wildlife are conservation priorities in many regions worldwide and the proposed approach may therefore be transferable to comparable open-habitat monitoring applications.  
Further validation across additional ecosystems and species groups is an important direction for future work on monocular 3D perception of wildlife from drone data.
We hope the benchmark helps connect ecological drone monitoring with progress in computer vision in monocular 3D scene understanding.

\section*{Acknowledgements}
This research has been carried out as part of the WildDrone project, funded by the European Union's Horizon Europe Research and Innovation Program under the Marie Skłodowska-Curie Grant Agreement No. 101071224. 
The WildDrone project has also received funding in part from the EPSRC-funded Autonomous Drones for Nature Conservation Missions grant (EP/X029077/1).
Field testing was carried out with permission from the Kenyan National Commission for Science, Technology and Innovation (NACOSTI/P/25/415376) and from the Kenyan Wildlife Research and Training Institute (WRTI/RPC/2025-384485).
The Grevy’s zebra data were collected at Mpala Conservancy in Kenya with permission from the Kenyan National Commission for Science, Technology and Innovation (NACOSTI/P/17/59088/15489, NACOSTI/P/59088/21567) and in affiliation with the Kenya Wildlife Service.
The authors would like to thank, in no particular order, for their support during the project: Kenya Wildlife Service, Ol Pejeta management, Kenya Flying Labs, Kenya Civil Aviation Authority, the local Kenya Air Force ATC, the Wildlife Research and Training Institute, and Mpala Conservancy.

\bibliographystyle{plainnat}
\bibliography{fin_citations}

\newpage

\appendix

\section{Supplementary Material}
\addcontentsline{toc}{section}{Supplementary Material}

\subsection{Dataset details}                      
\label{app:dataset}

\paragraph{Species and conservation status.}
\label{app:species}
Table~\ref{tab:species} lists the species annotated in WildBox with their
scientific names and IUCN Red List status (accessed 2026). The rhino
benchmark class combines black and white rhinoceros; all other classes
correspond to a single species.

\begin{table}[h]
\centering
\caption{Species annotated in WildBox and their IUCN Red List status.}
\label{tab:species}
\small
\begin{tabular}{@{}llll@{}}
\toprule
Benchmark class & Common name & Scientific name & IUCN status \\
\midrule
giraffe         & Reticulated giraffe   & \emph{Giraffa reticulata}     & Endangered \\
plains zebra    & plains zebra          & \emph{Equus quagga}           & Near Threatened \\
Gr\'evy's zebra & Gr\'evy's zebra       & \emph{Equus grevyi}           & Endangered \\
elephant        & African bush elephant & \emph{Loxodonta africana}     & Endangered \\
gazelle         & Thomson's gazelle     & \emph{Eudorcas thomsonii}     & Least Concern \\
\multirow{2}{*}{rhinoceros}
                & Black rhinoceros      & \emph{Diceros bicornis}       & Critically Endangered \\
                & White rhinoceros      & \emph{Ceratotherium simum}    & Near Threatened \\
\bottomrule
\end{tabular}
\end{table}

\paragraph{Per-class inventory.}
Table~\ref{tab:full_inventory} reports complete video, segment, frame,
and instance counts per class and per split.

\begin{table}[h]
\centering
\caption{Full per-class inventory across train and val splits. Rare
classes (gazelle, giraffe, Grévy's zebra) have $\le 4$ training videos
each; the remaining three classes have 12--15 training videos each.}
\label{tab:full_inventory}
\small
\begin{tabular}{lrrrrrrrr}
\toprule
Classes & tr.\ vid & val. vid & tr.\ seg & val. seg & tr.\ frm & val. frm & tr.\ inst & val. inst \\
\midrule
elephant         & 14 & 3 & 73 & 15 & 12{,}065 & 2{,}018 & 25{,}808 & 17{,}119 \\
gazelle          &  3 & 1 & 27 & 12 &  5{,}158 & 2{,}285 & 43{,}280 & 16{,}099 \\
giraffe          &  3 & 1 &  9 &  1 &  1{,}420 &    110 &  3{,}578 &      110 \\
Grévy's zebra    &  4 & 1 & 24 &  6 &  3{,}929 &    681 & 19{,}643 &  3{,}405 \\
plains zebra     & 12 & 3 & 58 & 20 & 10{,}186 & 3{,}725 & 49{,}864 & 12{,}586 \\
rhino            & 15 & 4 & 72 & 28 & 13{,}221 & 4{,}960 & 28{,}381 & 17{,}632 \\
\midrule
\textbf{Total}   & 51 &13 &263 & 82 & 45{,}979 &13{,}779 &170{,}554 & 66{,}951 \\
\bottomrule
\end{tabular}
\end{table}

\paragraph{Geometry and scene composition.}

\begin{figure}[ht]
\centering
\begin{subfigure}[t]{0.49\textwidth}
\centering
\includegraphics[width=\textwidth]{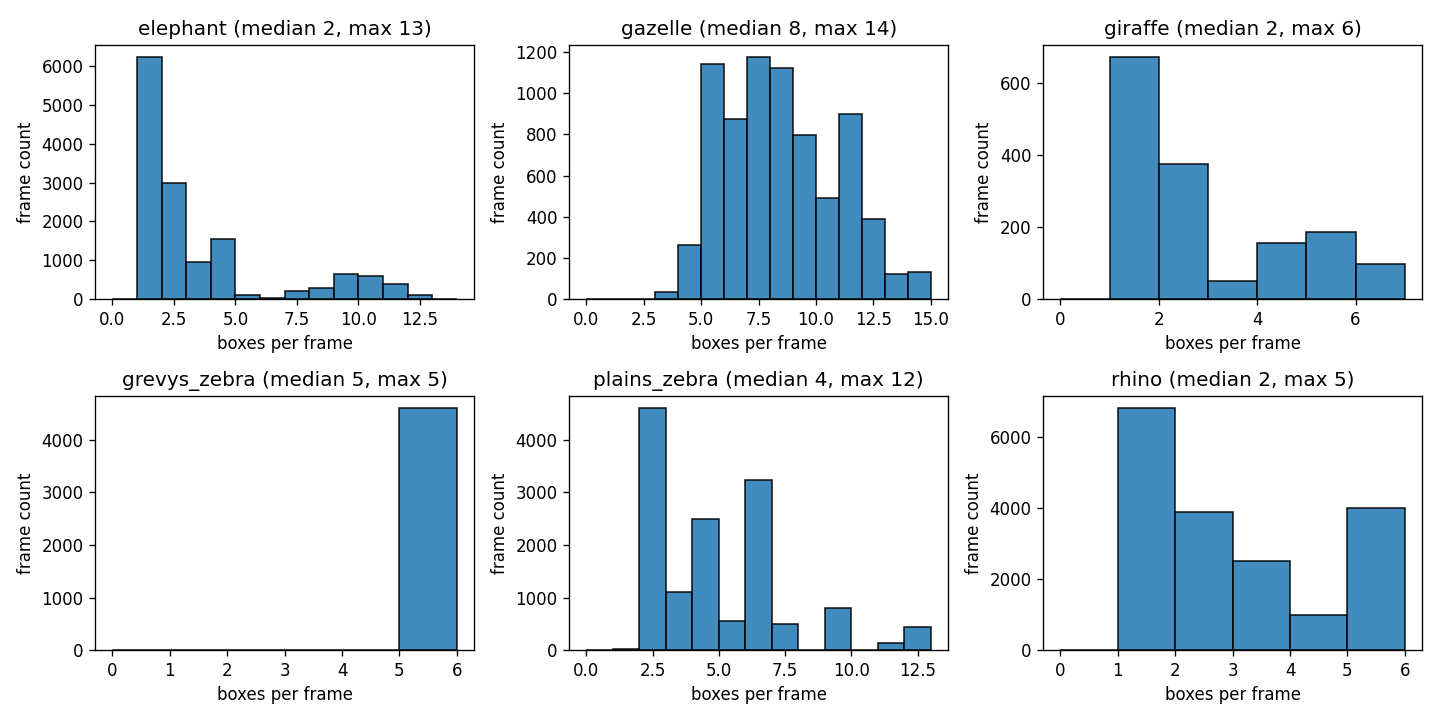}
\caption{Scene density (boxes per frame). Grévy's zebra is degenerate
at 5 boxes/frame; elephant is bimodal; plains zebra and rhino show
multi-modal structure.}
\label{fig:bbox_density}
\end{subfigure}
\hfill
\begin{subfigure}[t]{0.49\textwidth}
\centering
\includegraphics[width=\textwidth]{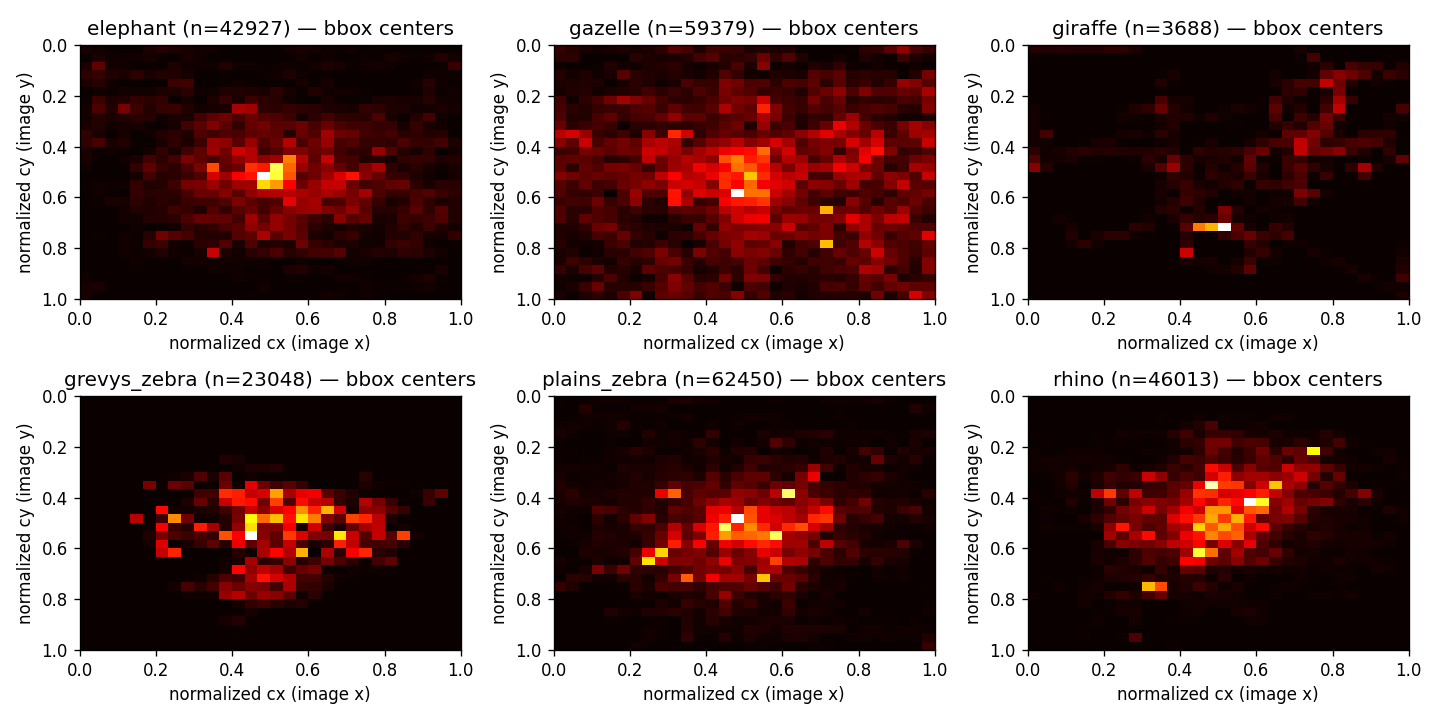}
\caption{Normalised bbox-centre position $(c_x, c_y)$. All classes are
centred near the image midpoint with a mild downward bias consistent
with oblique drone framing; the giraffe panel reflects sample-size
sparsity.}
\label{fig:bbox_position}
\end{subfigure}
\caption{Per-class distributions of scene density and bbox-centre
position, train+val combined. Per-class summary statistics are
reported in Table~\ref{tab:bbox_stats}.}
\label{fig:scene_composition}
\end{figure}
All annotated frames in WildBox are single-class: each video follows a
single herd, and no frame contains instances of more than one class.
Within-frame 2D class confusion is therefore structurally absent in this
release, and the benchmark isolates the 3D lifting problem from
multi-class classification ambiguity. Table~\ref{tab:bbox_stats}
summarises per-class 2D bounding-box geometry and scene density. Aerial
perspective produces two orders of magnitude spread in object size
across taxa, with gazelle the smallest (median 0.34\% of frame) and
elephant the largest (median 2.56\%). Aspect ratios cluster near 1.0
across classes; giraffe is the only sub-1 class (median 0.84) and
zebras the widest. Median scene density varies sharply with taxon, with
gazelle scenes roughly four times denser than the other classes.

\begin{table}[h]
\centering
\caption{Per-class 2D bounding-box geometry and scene density,
train+val combined. Quantiles reported as median [p25, p75]. All
images are 1920$\times$1080.}
\label{tab:bbox_stats}
\small
\begin{tabular}{lrrr}
\toprule
Classes & area (\%) & W/H & boxes/frame \\
\midrule
elephant         & 2.56 [1.37, 4.30] & 1.01 [0.67, 1.38] & 2 [1, 4]   \\
gazelle          & 0.34 [0.21, 0.49] & 1.07 [0.74, 1.41] & 8 [6, 10]  \\
giraffe          & 1.88 [0.99, 5.63] & 0.84 [0.59, 1.18] & 2 [1, 4]   \\
Grévy's zebra    & 0.59 [0.34, 0.89] & 1.25 [0.90, 1.52] & 5 [5, 5]   \\
plains zebra     & 1.42 [0.99, 1.99] & 1.29 [0.89, 1.56] & 4 [2, 6]   \\
rhino            & 1.62 [0.67, 3.74] & 1.09 [0.75, 1.58] & 2 [1, 4]   \\
\bottomrule
\end{tabular}
\end{table}

Figures~\ref{fig:bbox_density} and~\ref{fig:bbox_position} show the
full per-class distributions of scene density and bbox-center position.
The density histograms reveal structure not visible in the summary
quantiles: Grévy's zebra frames are degenerate at 5 boxes per frame,
elephant scenes are bimodal (a primary mode at 1--3 individuals and a
secondary mode at 8--11), and plains zebra and rhino show multi-modal
structure consistent with herds of varying size. The position
histograms show all classes centred near the image midpoint with the
moderate downward bias expected from oblique drone framing; the
giraffe panel reflects the limited sample size (n=3{,}688 boxes from a
single training video plus the validation video) rather than a
class-specific bias.

\paragraph{Track level statistics}
Table~\ref{tab:track_stats} tabulates the per-species track inventory.
WildBox contains 1{,}478 tracks (1{,}033 train / 445 val); a track spans 160.7 frames on average (median 200), and mean frames per track ranges from 131.7 (giraffe) to 177.7 (rhino). 
Because the same physical animal may be re-tracked in a later segment of the same encounter, the number of distinct individuals is bounded by $67 \le N_{\text{ind}} \le 1{,}478$, where the lower bound is the sum over the 11 shoots of the maximum number of animals co-visible in a single frame.

\begin{table}[h]
\centering
\caption{Instance-level statistics of WildBox. A track is one animal
instance followed through one segment, keyed
\texttt{(video,\,segment,\,track\_id)}; an annotation is one animal in
one frame. The released \texttt{track\_id} is numbered within a
segment, so tracks must be keyed jointly with video and segment.}
\label{tab:track_stats}
\small
\setlength{\tabcolsep}{4pt}
\begin{tabular}{l ccc ccc r cc}
\toprule
 & \multicolumn{3}{c}{Coverage} & \multicolumn{3}{c}{Tracks} & &
   \multicolumn{2}{c}{Frames / track} \\
\cmidrule(lr){2-4}\cmidrule(lr){5-7}\cmidrule(lr){9-10}
Species & Shoots & Videos & Segs. & All & Train & Val & Annotations & Mean & Median \\
\midrule
plains zebra    &  2 & 15 &  78 &  375 &  296 &  79 & 62{,}450 & 166.5 & 200.0 \\
gazelle         &  1 &  4 &  39 &  362 &  260 & 102 & 59{,}379 & 164.0 & 200.0 \\
elephant        &  3 & 17 &  88 &  304 &  169 & 135 & 42{,}927 & 141.2 & 179.5 \\
rhino           &  2 & 19 & 100 &  259 &  161 &  98 & 46{,}013 & 177.7 & 200.0 \\
Gr\'evy's zebra &  1 &  5 &  30 &  150 &  120 &  30 & 23{,}048 & 153.7 & 175.0 \\
giraffe         &  2 &  4 &  10 &   28 &   27 &   1 &  3{,}688 & 131.7 & 116.0 \\
\midrule
total           & 11 & 64 & 345 & 1{,}478 & 1{,}033 & 445 & 237{,}505 & 160.7 & 200.0 \\
\bottomrule
\end{tabular}
\end{table}

\paragraph{Validation-split sensitivity.}
The sensitivity of headline metrics to the val-split seed choice is
reported in Appendix~\ref{app:experiments}.

\subsection{Dataset labelling}

\subsubsection{Labelling interface: WildLIFT-A}
\label{app:anno_vis}

\begin{figure}[ht!]
    \centering
    \includegraphics[width=1\linewidth]{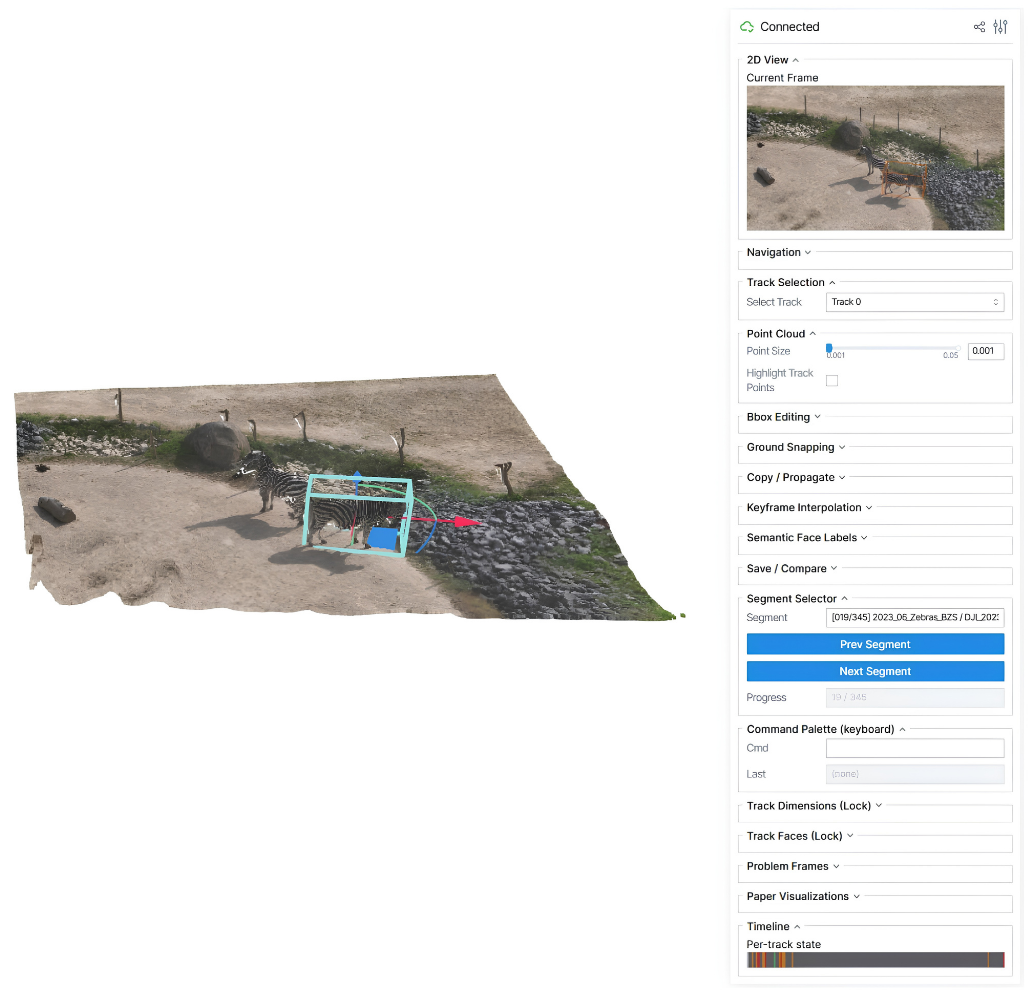}
    \caption{\textbf{Browser-based editor for refining automatically fitted 3D boxes.} Left: the 3D viewport shows the reconstructed pointmap for the current frame together with the oriented bounding box (OBB) of the selected track (cyan). The annotator can orbit, pan and zoom the scene and adjust the OBB directly in 3D via the attached gizmo, which provides translation along each axis (coloured arrows) and rotation (arc handle) for full 6-DOF correction. Right: the 2D view shows the source frame with the box projected into the image, with a toggle to the original automatic fit for comparison. Per track, the annotator can lock box dimensions to a consensus estimate across all frames, snap the box to the ground plane, assign semantic faces (front, top, left) once and propagate them, and interpolate between two keyframes. A problem-frame ranking based on dimension drift, velocity outliers and low confidence orders frames for review, and the timeline strip marks corrected, keyframe and flagged frames. A single-key command palette drives the whole workflow. Edits are written back as the tracking summary, KITTI-style labels and a per-session correction log.}
    \label{fig:geometry_label}
\end{figure}

\FloatBarrier

\subsubsection{Why WildBox uses video-based relative 3D annotation}
\label{app:annotation_scaling}

This appendix documents the requirements WildBox places on its
annotation pipeline and the measurements supporting our choice of
WildLIFT~\cite{shuklaWildLIFTLiftingMonocular2026}. The two closest
alternatives by output type are LabelAny3D~\cite{yao2025labelany3d},
which produces 3D bounding boxes from single images via analysis-by-synthesis,
and OVM3D-Det~\cite{huangTrainingOpenVocabularyMonocular2024b}, which
generates per-image pseudo-LiDAR boxes for open-vocabulary detector
training. Both operate at the single-image level. WildBox releases
\emph{tracked} per-segment 3D bounding boxes alongside cross-frame instance
identities (\S\ref{sec:dataset}), so a video-segment-level pipeline
that maintains identity across frames matches the release schema
directly, whereas single-image pipelines would require an additional
identity-association stage that is not part of either reference
method.

\paragraph{Throughput at release scale.}
We measured per-image runtime for LabelAny3D on a single NVIDIA A40
GPU, averaged over a batch of eight WildBox scenes.
Table~\ref{tab:labelany3d_runtime} reports the per-stage average. The
aggregate cost is approximately 306 seconds per image, of which scene
reconstruction accounts for the largest share. At WildBox's release
scale of 59{,}758 frames, this extrapolates to approximately
5{,}076 GPU-hours; on the same hardware, the WildLIFT pipeline
processes the full release in approximately 8 GPU-hours.

\begin{table}[h]
\centering
\caption{LabelAny3D per-stage runtime on a single NVIDIA A40 GPU,
averaged over a batch of eight WildBox scenes. Aggregate runtime is
approximately 306 seconds per image; reconstruction is the dominant
stage at $\sim$43\% of total time.}
\label{tab:labelany3d_runtime}
\small
\begin{tabular}{lrrrrrrrr}
\toprule
Stage & depth & enhance & crops & completion & elev & recon & whole & total \\
\midrule
Time (s) & $\sim$16 & $\sim$66 & $\sim$7 & $\sim$48 & $\sim$18 & $\sim$130 & $\sim$21 & $\sim$306 \\
\bottomrule
\end{tabular}
\end{table}

\paragraph{Scale regime.}
WildBox releases relative-scale 3D annotations
(\S\ref{sec:dataset}), so the annotation pipeline is not required to
recover metric scale. WildLIFT-A's per-segment relative reconstruction
matches this regime directly.

\subsubsection{Human review statistics}
\label{app:human_review}

All released 3D annotations underwent human review using the WildLIFT-A platform. 
The review was performed primarily at the tracklet level, allowing corrections to be propagated across the frames belonging to the same animal. 
WildLIFT-A records changes made during review, but it does not store a separate confirmation when an annotation is accepted without modification. 
We therefore quantified the review outcomes by matching each released annotation to its corresponding automatic annotation by (track, frame) and comparing its centre, dimensions and orientation. 
Unedited values are stored verbatim by the review tool, so exact numerical comparison identifies retained annotations; with a $10^{-12}$ tolerance the counts below shift by at most 0.12\%. Of the 237{,}532 reviewed annotations (237{,}505 after export filtering), 153{,}185 (64.5\%) were modified during human review, while 84{,}347 (35.5\%) were retained without numerical changes. 
No annotation was discarded during the human correction stage; annotations requiring refinement were corrected and retained. 
Only 3 of the 64 videos required no modification at all.

The corrections primarily affected object dimensions (Table~\ref{tab:human_review}). 
Of the modified annotations, 99.9\% had their box dimensions adjusted, whereas the centre changed in 0.2\% and the orientation in 1.4\%. 
Among the modified annotations, the median relative change of the box diagonal was 7.6\% (90th percentile: 26.8\%); centre shifts had a median magnitude of 0.12 box diagonals. 
Of the 2{,}106 orientation changes, 90\% corresponded to a signed axis permutation rather than changes in the estimated geometry; the remaining 203 genuine orientation corrections had a median angle of $84^\circ$, consistent with heading corrections.

The number of modified annotations should not be interpreted as the number of independent manual edits. Corrections were predominantly applied through track-level operations: of the 914 tracks with at least one modified annotation, 906 carry a single constant correction propagated across the track, accounting for 99.9\% of modified boxes, while 8 tracks (113 boxes) received frame-varying corrections. 
The 153{,}185 modified boxes therefore correspond to approximately 1{,}000 discrete reviewer actions.

\begin{table}[h]
\centering
\caption{Human-review outcome per species: released annotations
compared numerically to their automatic counterparts, matched by
(track, frame). Change-type columns count modified annotations whose
dimensions, centre or orientation differ; one annotation can appear
in several columns.}
\label{tab:human_review}
\small
\begin{tabular}{lrrrrrr}
\toprule
Species & Annotations & Modified & \% & Dims & Centre & Orient. \\
\midrule
plains zebra    & 62{,}454 & 59{,}143 & 94.7 & 59{,}042 &  49 & 1{,}923 \\
gazelle         & 59{,}399 &  3{,}521 &  5.9 &  3{,}521 &   0 &       0 \\
rhino           & 46{,}013 & 44{,}487 & 96.7 & 44{,}487 & 138 &     138 \\
elephant        & 42{,}930 & 20{,}439 & 47.6 & 20{,}439 &  45 &      45 \\
Gr\'evy's zebra & 23{,}048 & 22{,}361 & 97.0 & 22{,}361 &   0 &       0 \\
giraffe         &  3{,}688 &  3{,}234 & 87.7 &  3{,}234 &   0 &       0 \\
\midrule
total           & 237{,}532 & 153{,}185 & 64.5 & 153{,}084 & 232 & 2{,}106 \\
\bottomrule
\end{tabular}
\end{table}

\subsection{Extended experimental results}
\label{app:experiments}

This appendix reports per-class results for the headline fine-tuned
configurations, per-class evidence supporting the curriculum
attribution in \S\ref{sec:exp:curriculum}, the NHD metric definition,
and per-run hyperparameters.

\subsubsection{Per-class fine-tuned 3D AP}
\label{app:per_class_ft}

Table~\ref{tab:per_class_ft} reports per-class AP3D for the two
OVMono3D-LIFT fine-tuned configurations in
Table~\ref{tab:main_benchmark} and the DetAny3D 2-epoch fine-tuned
configuration; the per-class pattern supporting \S\ref{sec:exp:class}
is visible directly.

\begin{table}[h]
\centering
\caption{Per-class AP3D for fine-tuned configurations. OVMono3D-LIFT
rows are 3-seed mean\,$\pm$\,std. DetAny3D ep2 is reported from a
single seed.}
\label{tab:per_class_ft}
\small
\begin{tabular}{lrrr}
\toprule
Class & LIFT direct (A) & LIFT curriculum (D) & DetAny3D ep2 \\
\midrule
giraffe        & 10.67\,\scriptsize{$\pm$7.78} & 12.03\,\scriptsize{$\pm$3.91} & 14.53 \\
Gr\'evy's zebra & 6.51\,\scriptsize{$\pm$1.35} & 16.02\,\scriptsize{$\pm$1.12} &  0.00 \\
elephant       & 8.32\,\scriptsize{$\pm$0.37} &  9.62\,\scriptsize{$\pm$0.29} &  4.34 \\
plains zebra   &23.65\,\scriptsize{$\pm$2.84} & 33.40\,\scriptsize{$\pm$0.34} &  2.97 \\
rhino          & 4.10\,\scriptsize{$\pm$0.43} &  5.43\,\scriptsize{$\pm$0.29} &  2.02 \\
gazelle        & 1.47\,\scriptsize{$\pm$0.15} &  2.49\,\scriptsize{$\pm$0.22} &  1.05 \\
\midrule
macro          & 9.12\,\scriptsize{$\pm$0.84} & 13.17\,\scriptsize{$\pm$0.69} &  4.15 \\
\bottomrule
\end{tabular}
\end{table}

\subsubsection{DetAny3D epoch sweep}
\label{app:detany3d_epochs}

Table~\ref{tab:detany3d_epochs} supports the choice of two epochs for
DetAny3D fine-tuning (\S\ref{sec:eval}). One- and two-epoch
performance is comparable: AP3D differs by 0.4 (ep1 higher) and
BEV@0.50 by 0.07 (ep2 higher). We report two epochs as the headline
configuration on the basis of the BEV@0.50 advantage --- our primary
3D metric under the per-segment scale regime --- with one epoch a
defensible alternative. Three or more epochs drops both AP metrics
(AP3D $-0.4$, BEV@0.50 $-0.6$ relative to two epochs), indicating
overfitting.

\begin{table}[h]
\centering
\caption{DetAny3D fine-tuning at one, two, and three-or-more epochs on
WildBox validation (seed 0). Bold marks the recommended configuration.}
\label{tab:detany3d_epochs}
\small
\begin{tabular}{lrrrrr}
\toprule
Epochs & AP-BEV@0.50 & AP3D & Rel-AP3D & overall NHD & $z$/overall \\
\midrule
1   & 1.92 & 4.54 & 4.48 & 11.49 & 92.0\% \\
\textbf{2}   & \textbf{1.99} & 4.15 & 4.15 & 12.35 & 92.9\% \\
3+  & 1.42 & 3.72 & 3.73 & 11.40 & 92.2\% \\
\bottomrule
\end{tabular}
\end{table}

\subsubsection{Curriculum ablation: per-class evidence}
\label{app:curriculum_per_class}

Table~\ref{tab:per_class_ablation} reports per-class AP3D and
AP-BEV@0.50 for the three configurations in
Table~\ref{tab:ablations} that share the same sampler setting
(RT=0.5): direct fine-tuning at 15k iterations (A), direct
fine-tuning at 25k iterations (C), and curriculum-initialised
fine-tuning (D). The per-class evidence supports the attribution
in \S\ref{sec:exp:curriculum}: extra direct iterations do not
reproduce the curriculum gain.

\begin{table}[h]
\centering
\caption{Per-class AP3D and AP-BEV@0.50 for OVMono3D-LIFT direct
at 15k iterations (A), direct at 25k iterations (C), and
curriculum-initialised at 10k continuation (D). A and D are 3-seed
mean\,$\pm$\,std; C is seed 0.}
\label{tab:per_class_ablation}
\small
\setlength{\tabcolsep}{4pt}
\begin{tabular}{lrrrrrr}
\toprule
& \multicolumn{3}{c}{AP3D} & \multicolumn{3}{c}{AP-BEV@0.50} \\
\cmidrule(lr){2-4} \cmidrule(lr){5-7}
Class & A (15k) & C (25k) & D (10k cont.) & A (15k) & C (25k) & D (10k cont.) \\
\midrule
giraffe        & 10.67\,\scriptsize{$\pm$7.78} & 20.39 & 12.03\,\scriptsize{$\pm$3.91} & 0.15\,\scriptsize{$\pm$0.08} & 0.08 & 1.01\,\scriptsize{$\pm$1.43} \\
Gr\'evy's zebra & 6.51\,\scriptsize{$\pm$1.35} &  4.51 & 16.02\,\scriptsize{$\pm$1.12} & 7.81\,\scriptsize{$\pm$4.36} & 2.42 & 11.18\,\scriptsize{$\pm$2.82} \\
elephant       & 8.32\,\scriptsize{$\pm$0.37} &  8.55 &  9.62\,\scriptsize{$\pm$0.29} & 3.45\,\scriptsize{$\pm$0.21} & 3.67 & 10.22\,\scriptsize{$\pm$1.26} \\
plains zebra   &23.65\,\scriptsize{$\pm$2.84} & 25.70 & 33.40\,\scriptsize{$\pm$0.34} &14.76\,\scriptsize{$\pm$6.25} &16.86 & 22.50\,\scriptsize{$\pm$0.52} \\
rhino          & 4.10\,\scriptsize{$\pm$0.43} &  3.83 &  5.43\,\scriptsize{$\pm$0.29} & 1.47\,\scriptsize{$\pm$0.28} & 1.21 &  5.23\,\scriptsize{$\pm$3.93} \\
gazelle        & 1.47\,\scriptsize{$\pm$0.15} &  1.60 &  2.49\,\scriptsize{$\pm$0.22} & 0.75\,\scriptsize{$\pm$0.06} & 0.77 &  1.96\,\scriptsize{$\pm$1.10} \\
\midrule
macro          & 9.12\,\scriptsize{$\pm$0.84} & 10.76 & 13.17\,\scriptsize{$\pm$0.69} & 4.73\,\scriptsize{$\pm$1.80} & 4.17 &  8.68\,\scriptsize{$\pm$0.47} \\
\bottomrule
\end{tabular}
\end{table}

Three observations support the attribution. First, the per-class
rank ordering of A and C is identical (Spearman $\rho = 1.00$);
both match the per-class A--D and C--D orderings only weakly
($\rho = 0.83$). Extra direct iterations preserve the per-class
profile of A; curriculum-init reshapes it. Second, C's AP3D gain
over A concentrates on giraffe ($+9.7$); the remaining three
non-zebra classes are within $\pm 0.7$ AP3D of A. 
Curriculum-init instead gains $+9.5$ on Gr\'evy's zebra and $+9.8$ on plains zebra over A,
and exceeds C on every class on AP-BEV@0.50. Third, on
AP-BEV@0.50, extra direct iterations are net negative: C's macro
(4.17) is below A's (4.73). The curriculum gain is therefore not
explained by additional 6-class fine-tuning.

\subsubsection{NHD: definition}
\label{app:nhd_def}

For a matched (prediction, ground-truth) pair represented as 8
corner points $P, G \in \mathbb{R}^{8 \times 3}$, the per-pair
normalised Hausdorff distance is the optimal corner-to-corner
assignment cost, normalised by the GT cuboid diagonal:
\begin{equation*}
\mathrm{NHD}(P, G) =
  \frac{1}{\|G_{\max} - G_{\min}\|_2}
  \sum_{i=1}^{8} \|P_{\sigma^*(i)} - G_i\|_2,
\quad
\sigma^* = \arg\min_{\sigma \in S_8} \sum_{i=1}^{8} \|P_{\sigma(i)} - G_i\|_2,
\end{equation*}
with $\sigma^*$ obtained by Hungarian assignment over the 8 corners.
Pairs are matched by 2D IoU $\geq 0.5$, greedy per image. Every
component (xy, z, dimensions, pose) is normalised by the same GT
diagonal, making them directly comparable on a single scale.

The disentangled per-component NHD for component $c$ replaces all
components of $P$ except $c$ with the corresponding components of
$G$, yielding a modified prediction $P^{(c)}$, and evaluates
$\mathrm{NHD}(P^{(c)}, G)$. This isolates the corner error attributable
to component $c$ when every other component is held at GT.

We report the depth share of overall NHD as
$z\,/\,\mathrm{overall}$: the leave-one-in NHD for depth divided
by the unmodified overall NHD.

\subsubsection{Performance association with acquisition conditions}
\label{app:per_acqui}
We analysed how acquisition conditions are associated with both 2D and 3D performance of the headline fine-tuned OVMono3D-LIFT configuration (Table~\ref{tab:main_benchmark}). 
DJI SRT telemetry provides measured drone altitude (height above take-off), digital zoom and gimbal pitch ($-90^\circ$ = nadir; values closer to $0^\circ$ are more oblique), while herd density (animals per frame) and apparent
animal size are computed from the annotations. 
Apparent animal size is $\sqrt{w \cdot h}$ of the annotated 2D box in pixels (median 130\,px).
All 64 flights were matched to an SRT file, with telemetry available for 99.9\% of released frames overall (97.5\% in the least complete flight). 
Three DJI SRT layouts were parsed, corresponding to the three drone models used during acquisition. Zoom is absent from two layouts, so the digital zoom analysis uses only directly recorded values and no missing values are imputed. 
The observed ranges were 3.1--79.5\,m altitude, 1.0--4.0$\times$ zoom, $-82.8^\circ$ to $-0.1^\circ$ gimbal pitch and 1.0--11.7 animals per frame. 
The SRT files themselves are not released because they contain sensitive flight and location metadata, particularly for rhino sequences, for which precise locations could create a conservation and security risk.

The unit of analysis is the validation segment (82 segments nested in 13 flights). 
Predictions are matched one-to-one to ground truth in score order at 2D IoU $\geq 0.5$. 
2D performance is recall@0.5, the fraction of ground-truth boxes recovered by class-agnostic, unthresholded predictions. 
3D metrics are computed only for detections correctly matched in 2D, so they isolate 3D quality given a correct 2D detection: the per-segment median NHD (\S\ref{app:nhd_def}) and the fraction of matched detections with NHD $< 0.5$, i.e.\ a predicted cuboid within half a body diagonal of the ground truth, which we read as usable-3D performance. 
The 3D columns use the 81 segments with at least 20 matched pairs (2D recall uses seed~0; 3D metrics are the 3-seed mean). 
Table~\ref{tab:acq_conditions} reports tie-corrected Spearman rank correlations at two levels: across segments, and conservatively across the 13 independent flights (one point per flight). 
Digital zoom covers 48 segments in 8 flights.

\begin{table}[h]
\centering
\caption{Spearman rank correlations ($\rho$, with $p$ in parentheses)
between acquisition conditions and detection performance, at the segment and flight levels, with the latter providing a more conservative analysis. 
2D recall@0.5 measures the fraction of
ground-truth boxes recovered at IoU $\geq 0.5$; median NHD and the
fraction of matched detections with NHD $< 0.5$ are computed only over
one-to-one 2D-matched pairs only. Digital zoom is recorded in only one
of the three SRT layouts, resulting in a smaller sample size ($n$).}
\label{tab:acq_conditions}
\small
\begin{tabular}{lrccc}
\toprule
Variable & $n$ & 2D recall@0.5 & median NHD & frac.\ NHD$<$0.5 \\
\midrule
\multicolumn{5}{l}{\emph{Segment level}} \\
altitude (m)            & 82 & $+0.30$ (0.007)   & $+0.09$ (0.443)   & $-0.16$ (0.144) \\
digital zoom ($\times$) & 48 & $+0.48$ (0.001)   & $+0.35$ (0.016)   & $-0.41$ (0.004) \\
gimbal pitch ($^\circ$) & 82 & $+0.04$ (0.748)   & $+0.47$ ($<$0.001) & $-0.36$ (0.001) \\
herd density            & 82 & $-0.46$ ($<$0.001) & $+0.45$ ($<$0.001) & $-0.18$ (0.103) \\
animal size (px)        & 82 & $+0.50$ ($<$0.001) & $-0.35$ (0.001)   & $+0.14$ (0.210) \\
\midrule
\multicolumn{5}{l}{\emph{Flight level}} \\
altitude (m)            & 13 & $+0.41$ (0.165) & $+0.42$ (0.156) & $-0.66$ (0.013) \\
digital zoom ($\times$) &  8 & $+0.72$ (0.043) & $+0.20$ (0.629) & $-0.35$ (0.399) \\
gimbal pitch ($^\circ$) & 13 & $+0.08$ (0.788) & $+0.13$ (0.668) & $+0.07$ (0.831) \\
herd density            & 13 & $-0.59$ (0.035) & $+0.58$ (0.036) & $-0.09$ (0.773) \\
animal size (px)        & 13 & $+0.69$ (0.010) & $-0.40$ (0.174) & $+0.12$ (0.694) \\
\bottomrule
\end{tabular}
\end{table}

Altitude is not negatively associated with 2D recall ($\rho=+0.41$ at the flight level, $+0.30$ at the segment level), but higher altitude is associated with lower usable-3D performance (flight-level $\rho=-0.66$ for the fraction of matched detections with NHD $<0.5$). 
Digital zoom is positively associated with 2D recall (flight-level $\rho=+0.72$) but also with higher 3D NHD (segment-level $\rho=+0.35$), suggesting that zoom may help preserve visibility without fully recovering geometric detail.
More oblique gimbal pitch shows little association with 2D recall, but is associated with higher 3D NHD (segment-level $\rho=+0.47$), suggesting a specific degradation in 3D detection quality. 
Herd density affects both stages: 2D recall decreases (flight-level $\rho=-0.59$) and 3D NHD increases (segment-level $\rho=+0.45$, flight-level $+0.58$), consistent with inter-animal occlusion. 
Larger apparent animal size is associated with higher 2D recall (flight-level $\rho=+0.69$) and lower 3D NHD (segment-level $\rho=-0.35$). In the worst terciles of digital zoom and of gimbal pitch, the depth share of NHD (\S\ref{app:nhd_def}) rises to 0.78 and 0.77 respectively, against 0.23--0.42 in the best terciles, consistent with the depth-dominated error budget reported in the main paper.

We did not analyse lighting because the videos in many scenes record the camera's automatic exposure response rather than scene illumination, so treating them as lighting measurements would be misleading. 
Since the validation set contains 13 flights, segments are nested within flights and species is partly confounded with flight, these results are descriptive associations over the acquisition range represented in WildBox rather than isolated causal effects.

\subsubsection{Micro-level results}
\label{app:micro}

Macro AP remains the primary metric because it gives equal importance to every species, including rare classes; this appendix reports frequency-sensitive summaries that make the effect of class imbalance explicit.
For 2D, 3D and 3D-Rel, the evaluation outputs allow the mean of the per-class AP values weighted by the number of ground-truth instances in each class.
We refer to this as \emph{weighted AP}, rather than micro AP, because detections are not pooled into a single precision--recall curve.
For BEV, the evaluation output additionally provides a true pooled-PR micro AP.
The two estimators measure different quantities and are not mathematically interchangeable, so they are reported in separate columns of Table~\ref{tab:micro_macro}.
The validation split contains 17{,}632 rhino, 17{,}119 elephant, 16{,}099 gazelle, 12{,}586 plains-zebra, 3{,}405 Gr\'evy's-zebra and 110 giraffe instances.

\begin{table}[h]
\centering
\caption{Macro versus frequency-sensitive AP for the headline fine-tuned model (3-seed mean).
Weighted AP is the GT-support-weighted mean of per-class APs; pooled micro AP pools all detections into a single precision--recall curve and is available from the BEV evaluation only.}
\label{tab:micro_macro}
\small
\begin{tabular}{lrrr}
\toprule
Metric & Macro & Weighted & Pooled micro \\
\midrule
AP2D         & 39.51 & 46.73 & --- \\
AP3D         & 13.17 & 11.60 & --- \\
AP3D-Rel     & 12.89 & 11.34 & --- \\
AP-BEV@0.25  & 25.74 & ---   & 21.30 \\
AP-BEV@0.50  &  8.68 & ---   &  6.73 \\
\bottomrule
\end{tabular}
\end{table}

For 2D, weighted AP is higher than macro AP, indicating that classes with more validation instances generally achieve higher AP than classes with fewer instances.
For 3D and 3D-Rel, weighted AP is lower than macro AP, meaning that classes with more validation instances generally have lower AP.
For BEV, pooled micro AP is lower than macro AP at both IoU thresholds.
The pooled BEV micro AP varies across seeds (19.63, 19.63 and 24.65 at IoU 0.25; 4.42, 4.49 and 11.29 at IoU 0.50): because pooling combines detection scores from all classes into a single precision--recall curve, the result can be sensitive to differences in score calibration across classes.
We therefore report pooled micro AP alongside, rather than in place of, the per-class summaries.

\subsubsection{Reproducibility}
\label{app:repro}

Table~\ref{tab:hyperparams} reports per-run training hyperparameters. OVMono3D-LIFT 6-class runs share DINOv2 ViT-B/14 backbone, CLIP ViT-B/16 open-vocabulary head, RepeatFactorTrainingSampler with REPEAT\_SQRT, SGD, IMS\_PER\_BATCH=8 (the 5-class pretrained checkpoint used IMS\_PER\_BATCH=4), weight decay $10^{-4}$, and depth-loss weight 0.5$\times$ the other 3D components. DetAny3D runs share SAM ViT-H, DINOv2 ViT-L/14, and UniDepth backbones with the published DetAny3D AdamW configuration (LR $10^{-5}$; UniDepth $10^{-7}$), weight decay $10^{-7}$, fp32.

\begin{table}[h]
\centering
\caption{Per-run training hyperparameters. $t_R$ is REPEAT\_THRESHOLD
for class-balanced sampling. Curriculum-init (D) continues from a
10k-iteration five-class merged-zebra checkpoint trained from the
Omni3D checkpoint with BASE\_LR=$10^{-3}$, IMS\_PER\_BATCH=4,
RT=0.5.}
\label{tab:hyperparams}
\small
\setlength{\tabcolsep}{4pt}
\begin{tabular}{llrrrrrl}
\toprule
Run & Init & MAX\_ITER & STEPS & BASE\_LR & WARMUP & $t_R$ & Seeds \\
\midrule
\multicolumn{8}{l}{\emph{OVMono3D-LIFT}} \\
A: Direct, baseline    & Omni3D    & 15000 & [9000, 13500]  & $2{\times}10^{-3}$ & 500 & 0.5  & 0,1,2 \\
B: Direct, $t_R$ 0.35     & Omni3D    & 15000 & [9000, 13500]  & $2{\times}10^{-3}$ & 500 & 0.35 & 0     \\
C: Direct, extended    & Omni3D    & 25000 & [15000, 22500] & $2{\times}10^{-3}$ & 500 & 0.5  & 0     \\
D: Curriculum-init     & 5cl ckpt  & 10000 & [6000, 9000]   & $1{\times}10^{-3}$ & 200 & 0.5  & 0,1,2 \\
\midrule
\multicolumn{8}{l}{\emph{DetAny3D (epochs, not iterations)}} \\
ep1                    & DetAny3D ckpt & 1 epoch & --      & $10^{-5}$ & 0   & --   & 0     \\
ep2                    & DetAny3D ckpt & 2 epochs & --     & $10^{-5}$ & 0   & --   & 0     \\
ep3+                   & DetAny3D ckpt & 3+ epochs & --    & $10^{-5}$ & 0   & --   & 0     \\
\bottomrule
\end{tabular}
\end{table}

All training and evaluation runs use 1$\times$ NVIDIA A40 (48\,GB)
GPUs with PyTorch 2.x (CUDA 11.8) and Python 3.8; OVMono3D-LIFT
relies on a pinned Detectron2 fork and a CPU-only pytorch3d build.
Wall-clock training time is approximately 3.5\,h for 10k-iteration
OVMono3D-LIFT runs, 9\,h for the 25k-iteration run, and 6\,h per
DetAny3D epoch. 
The dataset and baseline checkpoints are at
\url{https://huggingface.co/datasets/3DOMFBK/WildBox}.

\subsection{Qualitative results}
\label{app:qualitative}

\begin{figure}[ht!]
    \centering
    \includegraphics[width=\linewidth]{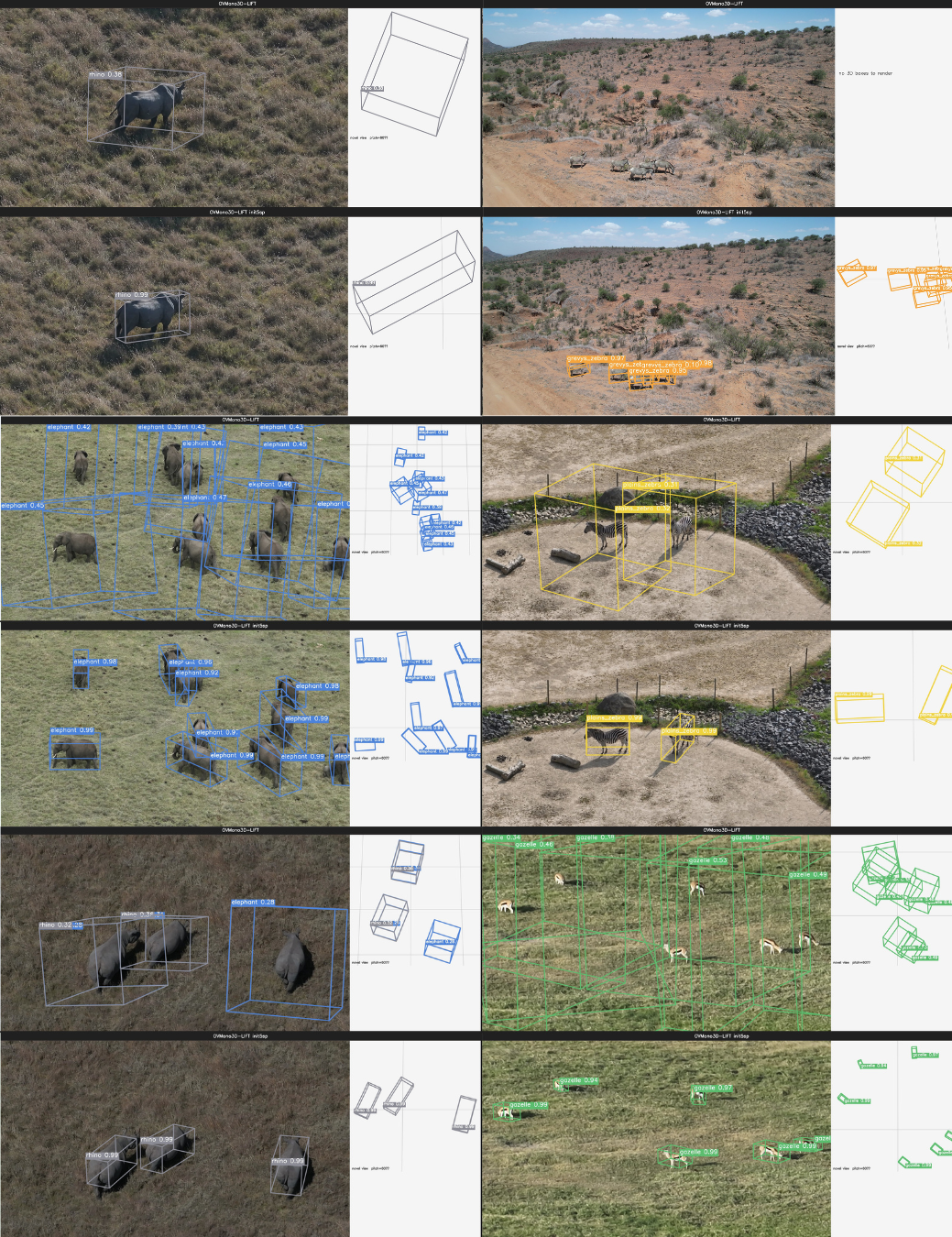}
    \caption{Qualitative comparison of OVMono3D-LIFT under
    ground-truth 2D box prompted zero-shot inference (odd rows) and
    supervised curriculum-initialised fine-tuning on WildBox (even
    rows). Six panels span five classes: rhino (rows~1--2,
    top-left), Grévy's zebra (rows~1--2, top-right), elephant
    (rows~3--4, left), plains zebra (rows~3--4, right), a second
    rhino sequence (rows~5--6, bottom-left), and gazelle
    (rows~5--6, bottom-right). Each panel pairs the camera-view
    projection of the predicted 3D boxes with a novel-view rendering.}
    \label{fig:qual}
\end{figure}

Figure~\ref{fig:qual} shows qualitative comparisons between
OVMono3D-LIFT under ground-truth 2D box prompted zero-shot inference and
the same architecture after curriculum-initialised fine-tuning on
WildBox. The visual pattern matches the disentangled NHD analysis
in~\S\ref{sec:exp:depth}: under zero-shot inference, the 3D head
produces inflated, near-uniform cuboids that ignore class scale and
inter-instance depth ordering, consistent with the 99.3\% depth share
of overall NHD reported in Table~\ref{tab:nhd} for the GT-2D-prompt
zero-shot condition. Fine-tuning on WildBox restores class-appropriate
extent, depth placement, and inter-instance geometry, consistent with
the recovery from 0.00 to 13.17 macro AP3D in
Table~\ref{tab:main_benchmark}.

\clearpage
\section*{NeurIPS Paper Checklist}

\begin{enumerate}

\item {\bf Claims}
    \item[] Question: Do the main claims made in the abstract and introduction accurately reflect the paper's contributions and scope?
    \item[] Answer: \answerYes{}
    \item[] Justification: The abstract and \S\ref{sec:intro} state the contributions (a 237{,}505-instance aerial monocular 3D detection dataset, baseline benchmarks for two open-vocabulary 3D architectures under zero-shot and fine-tuned settings, and a disentangled error analysis identifying depth as the dominant failure mode), and these match the experimental results
    reported in \S\ref{sec:experiments}.

\item {\bf Limitations}
    \item[] Question: Does the paper discuss the limitations of the work performed by the authors?
    \item[] Answer: \answerYes{}
    \item[] Justification: \S\ref{sec:limitations} addresses the limitations in detail.

\item {\bf Theory assumptions and proofs}
    \item[] Question: For each theoretical result, does the paper provide the full set of assumptions and a complete (and correct) proof?
    \item[] Answer: \answerNA{}
    \item[] Justification: The paper does not include theoretical results.

\item {\bf Experimental result reproducibility}
    \item[] Question: Does the paper fully disclose all the information needed to reproduce the main experimental results of the paper to the extent that it affects the main claims and/or conclusions of the paper (regardless of whether the code and data are provided or not)?
    \item[] Answer: \answerYes{}
    \item[] Justification: Training hyperparameters are reported in
    Appendix~\ref{app:repro}; metric definitions in \S\ref{sec:eval} and
    Appendix~\ref{app:nhd_def}. Code, dataset, and checkpoints are
    released as described in Q5.

\item {\bf Open access to data and code}
    \item[] Question: Does the paper provide open access to the data and code, with sufficient instructions to faithfully reproduce the main experimental results, as described in supplemental material?
    \item[] Answer: \answerYes{}
    \item[] Justification: The dataset is documented through per-class inventory tables (Appendix~\ref{app:dataset}), the annotation pipeline (\S\ref{sec:dataset}, Appendix~\ref{app:annotation_scaling}), Croissant 1.0 metadata in the OpenReview submission, and the training/evaluation code referenced in Appendix~\ref{app:repro}.

\item {\bf Experimental setting/details}
    \item[] Question: Does the paper specify all the training and test details (e.g., data splits, hyperparameters, how they were chosen, type of optimizer) necessary to understand the results?
    \item[] Answer: \answerYes{}
    \item[] Justification: \S\ref{sec:dataset} describes splits and the scale regime; \S\ref{sec:eval} specifies evaluation settings, training configurations, and metrics; per-run hyperparameters are in Appendix~\ref{app:repro}.
    
\item {\bf Experiment statistical significance}
    \item[] Question: Does the paper report error bars suitably and correctly defined or other appropriate information about the statistical significance of the experiments?
    \item[] Answer: \answerYes{}
    \item[] Justification: For the two headline OVMono3D-LIFT fine-tuned configurations (direct at 15k iterations and curriculum-init at 10k continuation) we report 3-seed mean$\pm$std on every metric in Table~\ref{tab:main_benchmark}; standard deviations capture
    seed-to-seed variability under fixed data and hyperparameters.
    DetAny3D fine-tuned numbers are reported from a single seed.

\item {\bf Experiments compute resources}
    \item[] Question: For each experiment, does the paper provide sufficient information on the computer resources (type of compute workers, memory, time of execution) needed to reproduce the experiments?
    \item[] Answer: \answerYes{}
    \item[] Justification: All training and evaluation runs use 1$\times$ NVIDIA A40 (48\,GB) GPUs. The details of the cluster are not disclosed to maintain anonymity and will be made available upon acceptance.
    
\item {\bf Code of ethics}
    \item[] Question: Does the research conducted in the paper conform, in every respect, with the NeurIPS Code of Ethics?
    \item[] Answer: \answerYes{}
    \item[] Justification: Drone footage was collected following applicable local laws and under valid permits, with efforts made to minimise disturbance to wildlife (\S\ref{sec:dataset}); the dataset contains only wildlife imagery and no human subjects.

\item {\bf Broader impacts}
    \item[] Question: Does the paper discuss both potential positive societal impacts and negative societal impacts of the work performed?
    \item[] Answer: \answerYes{}
    \item[] Justification: WildBox is intended to support wildlife monitoring, conservation research, and behavioural ecology, as motivated in \S\ref{sec:intro}. The dataset includes endangered species, so we treat poaching-related misuse as the principal negative-impact pathway and address it under safeguards (Q11).

\item {\bf Safeguards}
    \item[] Question: Does the paper describe safeguards that have been put in place for responsible release of data or models that have a high risk for misuse (e.g., pre-trained language models, image generators, or scraped datasets)?
    \item[] Answer: \answerYes{}
    \item[] Justification: No drone telemetry or other sensitive files are released for this dataset. 

\item {\bf Licenses for existing assets}
    \item[] Question: Are the creators or original owners of assets (e.g., code, data, models), used in the paper, properly credited and are the license and terms of use explicitly mentioned and properly respected?
    \item[] Answer: \answerYes{}
    \item[] Justification: All third-party assets are cited at point of use with supporting
    code dependencies in Appendix~\ref{app:repro}. WildBox itself will
    be released under CC-BY 4.0 upon acceptance.

\item {\bf New assets}
    \item[] Question: Are new assets introduced in the paper well documented and is the documentation provided alongside the assets?
    \item[] Answer: \answerYes{}
    \item[] Justification: The dataset is documented through per-class
    inventory tables (Appendix~\ref{app:dataset}), the annotation
    pipeline (\S\ref{sec:dataset}, Appendix~\ref{app:annotation_scaling}),
    Croissant 1.0 metadata in the open review submission, and an
    evaluation script that regenerates every headline number from
    saved predictions (Appendix~\ref{app:repro}).

\item {\bf Crowdsourcing and research with human subjects}
    \item[] Question: For crowdsourcing experiments and research with human subjects, does the paper include the full text of instructions given to participants and screenshots, if applicable, as well as details about compensation (if any)?
    \item[] Answer: \answerNA{}
    \item[] Justification: No crowdsourcing or human-subjects were involved in the research. 
    All annotations were reviewed by the authors.

\item {\bf Institutional review board (IRB) approvals or equivalent for research with human subjects}
    \item[] Question: Does the paper describe potential risks incurred by study participants, whether such risks were disclosed to the subjects, and whether Institutional Review Board (IRB) approvals (or an equivalent approval/review based on the requirements of your country or institution) were obtained?
    \item[] Answer: \answerNA{}
    \item[] Justification: No human subjects.

\item {\bf Declaration of LLM usage}
    \item[] Question: Does the paper describe the usage of LLMs if it is an important, original, or non-standard component of the core methods in this research? Note that if the LLM is used only for writing, editing, or formatting purposes and does \emph{not} impact the core methodology, scientific rigor, or originality of the research, declaration is not required.
    \item[] Answer: \answerNA{}
    \item[] Justification: LLM did not impact core methodology.

\end{enumerate}

\end{document}